\documentclass[twocolumn, switch]{article} 

\usepackage{titlesec}
\titleformat{\section}{\Large\bfseries\scshape}{\thesection}{1em}{}
\titleformat{\subsection}{\large\bfseries}{\thesubsection}{1em}{}
\titleformat{\subsubsection}{\normalsize\bfseries\itshape}{\thesubsubsection}{1em}{}
\titleformat{\paragraph}{\normalsize\bfseries}{\theparagraph}{1em}{}
\titlespacing\section{0pt}{12pt plus 3pt minus 3pt}{1pt plus 1pt minus 1pt}
\titlespacing\subsection{0pt}{10pt plus 3pt minus 3pt}{1pt plus 1pt minus 1pt}
\titlespacing\subsubsection{0pt}{8pt plus 3pt minus 3pt}{1pt plus 1pt minus 1pt}
\titlespacing\paragraph{0pt}{6pt plus 3pt minus 3pt}{1pt plus 1pt minus 1pt}
\newcounter{fourlevel}[subsubsection]
\renewcommand{\thefourlevel}{\thesubsubsection.\arabic{fourlevel}}
\newcommand{\fourlevelsection}[1]{%
  \refstepcounter{fourlevel}%
  \par\addvspace{6pt plus 3pt minus 3pt}%
  \noindent\textbf{\thefourlevel\quad #1}\par\nobreak\addvspace{1pt plus 1pt minus 1pt}%
  \addcontentsline{toc}{paragraph}{\protect\numberline{\thefourlevel}#1}%
}

\usepackage{preprint}
\makeatletter
\renewcommand{\section}{%
  \@startsection{section}{1}{\z@}%
                {-2.0ex \@plus -0.5ex \@minus -0.2ex}%
                { 1.5ex \@plus  0.3ex \@minus  0.2ex}%
                {\Large\bfseries\scshape\raggedright}%
}
\renewcommand{\subsection}{%
  \@startsection{subsection}{2}{\z@}%
                {-1.8ex \@plus -0.5ex \@minus -0.2ex}%
                { 0.8ex \@plus  0.2ex}%
                {\large\bfseries\raggedright}%
}
\renewcommand{\subsubsection}{%
  \@startsection{subsubsection}{3}{\z@}%
                {-1.5ex \@plus -0.5ex \@minus -0.2ex}%
                { 0.5ex \@plus  0.2ex}%
                {\normalsize\bfseries\itshape\raggedright}%
}
\makeatother
\usepackage[ruled,vlined,linesnumbered]{algorithm2e}
\usepackage{amsmath, amsthm, amssymb, amsfonts}

\usepackage[backend=biber, style=apa, natbib=true, doi=true, url=false, isbn=false]{biblatex}
\AtEveryBibitem{
  \clearfield{note}
  \clearfield{file}
  \clearfield{abstract}
  \clearfield{keywords}
  \clearfield{urldate}
  \clearfield{shorttitle}
  \clearfield{language}
  \clearfield{copyright}
  \clearfield{issn}
}

\usepackage{xcolor}		
\usepackage[colorlinks = true,
            linkcolor = purple,
            urlcolor  = blue,
            citecolor = cyan,
            anchorcolor = black]{hyperref}	
\usepackage{booktabs}
\usepackage{multirow} 		
\usepackage{nicefrac}		
\usepackage{microtype}		
\usepackage{float}
\usepackage{graphicx}
\usepackage[export]{adjustbox}
\usepackage[most]{tcolorbox}
\definecolor{mygray}{RGB}{240,240,240}
\tcbset{colback=mygray,boxrule=0pt,}
\graphicspath{ {./images/} }

\usepackage{newfloat}
\DeclareFloatingEnvironment[name={Supplementary Figure}]{suppfigure}
\usepackage{sidecap}
\sidecaptionvpos{figure}{c}

\title{Robust Dempster-Shafer Evidence Fusion with Chaos-Conflict Measurement and Historical-Experience Weighting}
\usepackage{titling}
\usepackage{orcidlink}
\usepackage{footmisc}
\newcommand{\Author}[3]{
  #1\textsuperscript{#2}%
  \ifstrempty{#3}{}{\,\orcidlink{#3}} %
}

\author{
  \Author{Huiyu Li}{1,2}{0000-0001-7291-490X} \and
  \Author{Weibo Liu}{2}{} \and
  \Author{Xinru Xu}{2}{} \and
  \Author{Dongchen Gao}{2}{} \and
  \Author{Meng Zhang}{2}{} \and
  \Author{Junhua Hu}{1}{}
}

\date{%
  \textsuperscript{1}School of Business, Central South University, Changsha, Hunan, People's Republic of China, 410083\\
  \textsuperscript{2}School of Management, Shandong University, Jinan, Shandong, People's Republic of China, 250100\\[1em]
  \footnotesize Corresponding author: Junhua Hu (\texttt{hujunhua@csu.edu.cn})\\
}

\newtheoremstyle{myplain}%
  {6pt}{6pt}{\itshape}{}{\bfseries}{.}{5pt}{}
\newtheoremstyle{mydef}%
  {6pt}{6pt}{}{}{\bfseries}{.}{5pt}{}
\newtheoremstyle{myremark}%
  {6pt}{6pt}{\itshape}{}{\bfseries\itshape}{.}{5pt}{}

\theoremstyle{myplain}

\newtheorem{lemma}{Lemma}
\newtheorem{corollary}{Corollary}

\theoremstyle{mydef}
\newtheorem{definition}{Definition}
\newtheorem{property}{Property}

\theoremstyle{myremark}
\newtheorem{remark}{Remark}
\theoremstyle{mydef}
\newtheorem{example}{Example}

\theoremstyle{mydef}

\theoremstyle{myremark}

\begin{document}
\hyphenpenalty=10000
\exhyphenpenalty=10000
\fancyhead{}
\renewcommand{\headrulewidth}{0pt}
\setcounter{MaxMatrixCols}{256}

\twocolumn[ 
  \begin{@twocolumnfalse} 

\maketitle
\thispagestyle{empty}

\begin{abstract}
Multi-source evidence fusion under Dempster-Shafer theory faces two persistent challenges: existing conflict measures assess inter-evidence inconsistency and intra-evidence uncertainty independently, yielding incomplete evaluations, and current fusion methods evaluate evidence sources exclusively through instantaneous comparisons without exploiting their long-term reliability across diverse decision contexts. This paper proposes a unified evidence reasoning framework that addresses both limitations. Specifically, a chaos-conflict measurement is introduced to jointly quantify cross-evidence conflict and intra-evidence non-specificity, with five formally proven properties ensuring consistent assessment. A historical experience driven weighting scheme partitions the decision space via spectral clustering and applies regret theory to compute context-specific reliability profiles from past fusion outcomes. These mechanisms feed into a hybrid combination rule that adaptively balances uncertainty preservation against weighted consensus, controlled by the global conflict level, followed by a belief-interval decision strategy that enables robust classification without discarding epistemic uncertainty. Experiments on 16 real-world benchmark datasets demonstrate that the proposed framework achieves an average F1 score of 85.78 and a mean AUC of 93.30, outperforming eight DST-based baselines and three gradient boosting methods. Ablation analysis confirms the contribution of each component we proposed. The framework offers an effective approach for adaptive evidence fusion in multi-source decision making.
\end{abstract}
\vspace{0.35cm}

  \end{@twocolumnfalse} 
] 


\section{Introduction}\label{sec:intro}

Multi-source information fusion plays a critical role in modern decision-making systems, where heterogeneous evidence must be integrated to support reliable judgments under uncertainty \citep{Xiao_2026,El_Din_2024}. Applications ranging from industrial fault diagnosis to pattern recognition and environmental monitoring share a common challenge: multiple evidence sources often provide assessments that are incomplete, imprecise, and mutually conflicting \citep{Yu_2026,Qiao_2023a,Li_2026}. The ability to fuse such evidence while preserving uncertainty information remains a fundamental problem in the fields of information systems and decision science.

Dempster-Shafer theory (DST) provides a generalization of Bayesian probability that accommodates partial belief and ignorance through the assignment of basic probability assignments (BPAs) to subsets of a hypothesis space \citep{Radzvilas_2024}. Unlike classical probability, DST assigns mass to multi-element subsets, thereby explicitly representing epistemic uncertainty. Despite its theoretical elegance, DST faces well-documented challenges when evidence sources exhibit high conflict \citep{Calderwood_2016,Zhou_2026}. The Zadeh paradox \citep{zadeh_simple_1986} demonstrated that Dempster's rule can produce counterintuitive results when two sources strongly disagree, and this limitation has motivated decades of research into conflict-aware fusion strategies.

Two broad categories of solutions have emerged. The first modifies the combination rule itself: \citet{yager_dempster-shafer_1987} reassigns conflict mass to the frame of discernment, \citet{dubois_representation_1988} propose a disjunctive rule that preserves uncertainty rather than normalizing it away, and various authors have introduced weighted, cautious, or transferable belief models to handle specific failure modes. The second category addresses the evidence sources directly, applying distance-based or similarity-based weighting schemes before fusion. \citet{murphy_combining_2000} averages BPAs uniformly; \citet{yong_combining_2004} weights sources using Jousselme distance; subsequent methods incorporate entropy, trust, or reliability coefficients to discount suspect evidence. Recent studies have attempted to combine preprocessing with adaptive fusion strategies. While these approaches have achieved incremental improvements, two persistent limitations remain. First, existing conflict measures \citep{destercke_toward_2013} treat uncertainty and conflict as independent dimensions, yielding assessments that are either incomplete or redundant. Second, all of the above methods evaluate evidence solely on the basis of instantaneous BPA comparisons, ignoring the long-term track record of evidence sources across diverse decision contexts \citep{Xu_2016}.

These limitations are consequential in practice. Existing models predict that evidence sources operating under similar conditions should exhibit consistent reliability \citep{Huang_2025}, yet empirical observation reveals the opposite: a source that produces misleading assessments in one scenario may be highly dependable in another, depending on the characteristics of the specific decision context \citep{Gao_2025}. Methods that assign static weights or that evaluate evidence exclusively through current pairwise comparisons cannot capture such context-dependent reliability \citep{Park_2025}, leading to two failure modes. Informative sources that are occasionally conflicting but generally dependable are indiscriminately suppressed, while systematically unreliable sources receive insufficient penalization \citep{Zhang_2025,Ghorbanzadeh_2021}. Furthermore, conventional conflict measures such as Dempster's conflict coefficient K consider only the total mass assigned to empty intersections \citep{Urbani_2023}, disregarding the internal structure of multi-element focal elements. This simplification leads to coarse assessments of evidence quality, particularly when sources allocate substantial mass to composite hypotheses that express genuine ignorance rather than true disagreement.

This paper proposes a unified evidence reasoning framework that addresses both limitations through two complementary mechanisms. The first is a chaos-conflict measurement (CCM) that jointly evaluates cross-evidence association and intra-evidence non-specificity within a single scalar quantity. The CCM is grounded in a novel evidence similarity measure whose properties of boundedness, symmetry, monotonicity, extreme consistency, and refinement insensitivity are formally proven. The second mechanism is a historical-experience-driven evidence weighting scheme that exploits past decision outcomes to learn context-dependent reliability profiles. Spectral clustering partitions the decision space into heterogeneous contexts, and regret theory is employed to quantify, for each evidence source, the counterfactual contribution when a fusion decision deviates from ground truth. Aggregated regret and rejoice scores are normalized to produce context-specific weights that reflect long-term source reliability. These two mechanisms feed into a hybrid combination rule that adaptively balances a conservative uncertainty preservation term against a weighted consensus component, with the tradeoff controlled by the global conflict level. A belief-interval decision rule then scores each hypothesis by combining its belief lower bound with a stability-weighted plausibility upper bound, enabling robust classification without forcing artificial redistribution of mass from multi-element focal elements.

The proposed framework is evaluated on 16 real-world benchmark datasets spanning biology, medicine, geography, and demography, using decision trees as evidence sources. Compared against eight DST-based fusion methods and three gradient boosting baselines, the framework achieves the best average rank across all evaluation metrics. Robustness analyses demonstrate stable performance under feature noise, label noise, hybrid noise scenarios, varying numbers of evidence sources, and substitution of the base evidence generator. Ablation experiments confirm that historical experience weighting provides the largest individual contribution to performance, reducing AUC by 5.03\% when removed. These results establish the proposed framework as an effective approach for evidence fusion that balances uncertainty preservation with adaptive reliability assessment, with direct applicability to any multi-source decision setting where evidence sources operate repeatedly across heterogeneous contexts.

The remainder of this paper is organized as follows. Section~\ref{sec:2_preliminaries} reviews the preliminary concepts of Dempster-Shafer theory, regret theory, and spectral clustering. Section~\ref{sec:3_proposed_approach} presents the proposed framework, including the chaos-conflict measurement, the historical experience driven weighting mechanism, the hybrid combination rule, and the belief-interval decision rule. Section~\ref{sec:4_algorithmic_design} details the algorithmic design of the offline training and online inference phases. Section~\ref{sec:5_experiment} reports the experimental setup, comparative results, robustness analyses, parameter sensitivity, and ablation study. Section~\ref{sec:6_conclusion} concludes the paper with a summary of contributions, a discussion of limitations, and directions for future research.

\section{Preliminaries}\label{sec:2_preliminaries}

\subsection{Dempster-Shafer theory}

DST, also known as belief function theory, is a mathematical framework for representing and reasoning with uncertain, imprecise, and incomplete information. Originating from the work of \citet{dempster_upper_1967} and formalized by \citet{shafer_mathematical_1976}, DST extends classical probability theory by allowing probability masses to be assigned not only to singletons but also to subsets of the hypothesis space. This provides a richer and more flexible representation of uncertainty, especially when prior knowledge is limited or heterogeneous evidence sources yield conflicting assessments.

\begin{definition} Frame of Discernment (FoD). Let \(\Theta  = \left\{  {{\theta }_{1},{\theta }_{2},\cdots ,{\theta }_{i},\cdots ,{\theta }_{n}}\right\}\) denote the frame of discernment, i.e., a set of mutually exclusive and collectively exhaustive hypotheses. DST operates on the power set: \label{def1} \end{definition}

\begin{multline}\tag{1} \label{eq:1} {2}^{ \Theta  } = \Bigl\{\varnothing,\; \{\theta_1\},\; \{\theta_2\},\; \{\theta_3\},\;\ldots,\\
\{\theta_1,\theta_2\},\; \{\theta_1,\theta_3\},\;\ldots,\; \{\theta_1,\theta_2,\theta_3\},\;\ldots,\; \Theta\Bigr\} \end{multline}
where each subset represents a proposition whose truth is uncertain. Here, \(\varnothing\) is an empty set. If \(S \in  {2}^{\Theta },S\) is called a hypothesis.

\begin{definition} Basic Probability Assignment. A BPA or mass function \(m : {2}^{\Theta } \rightarrow  \left\lbrack  {0,1}\right\rbrack\) satisfies: \label{def2} \end{definition}

\begin{equation}\tag{2} \label{eq:2} \left\{  \begin{array}{l} \mathop{\sum }\limits_{{S \subseteq  {2}^{\Theta }}}m\left( S\right)  = 1 \\  m\left( \varnothing \right)  = 0 \end{array}\right. \end{equation}

A subset \(S\) with \(m\left( S\right)  > 0\) is called a focal element (FE). Mass assigned to multi-element sets (N-Focal Elements, NFEs) expresses epistemic uncertainty or ignorance, since it indicates that the exact hypothesis cannot be precisely identified. Conversely, single-element focal elements (SFEs) represent precise hypotheses.

\begin{definition} Belief and plausibility functions. Based on BPA, DST defines two dual measures that jointly describe the degree of support for a proposition. \label{def3} \end{definition}

The belief function, representing the minimal support for \(S\) :

\begin{equation}\tag{3} \label{eq:3} \operatorname{Bel}\left( S\right)  = \mathop{\sum }\limits_{{A \subseteq  S}}m\left( A\right) \end{equation}

The plausibility function, representing the maximal potential support for \(S\) :

\begin{equation}\tag{4} \label{eq:4} {Pl}\left( S\right)  = 1 - {Bel}\left( \bar{S}\right)  = \mathop{\sum }\limits_{{A \cap  S \neq  \varnothing }}m\left( A\right) ,A \subseteq  \Theta \end{equation}

Together, \(\left\lbrack  {\operatorname{Bel}\left( S\right) ,\operatorname{Pl}\left( S\right) }\right\rbrack\) forms the belief interval, which provides a natural expression of uncertainty. A wider interval indicates greater ambiguity, while a narrow interval signifies strong confidence.

\begin{definition} Dempster's combination rule. One of DST's core strengths is its ability to integrate independent pieces of evidence. Given \(n\) mass functions \({m}_{1},{m}_{2},\ldots ,{m}_{n}\) defined on the same frame, Dempster's rule combines them as \label{def4} \end{definition}

\begin{equation}\tag{5} \label{eq:5} \left\{  \begin{array}{l} {m}_{1 \oplus  2 \oplus  \ldots  \oplus  n}\left( S\right)  = \frac{\mathop{\sum }\limits_{{\cap {A}_{i} = S,{A}_{i} \subseteq  {2}^{\Theta }}}\mathop{\prod }\limits_{j}{m}_{j}\left( {A}_{i}\right) }{1 - K} \\  K = \mathop{\sum }\limits_{{\bigcap {A}_{i} = \varnothing ,{A}_{i} \subseteq  {2}^{\Theta }}}{m}_{1}\left( {A}_{1}\right)  \cdot  {m}_{2}\left( {A}_{2}\right) \cdots {m}_{n}\left( {A}_{n}\right)  \end{array}\right. \end{equation}
where \(\oplus\) denotes Dempster's orthogonal sum, which aggregates independent pieces of evidence through combination. \(K \in  \left\lbrack  {0,1}\right\rbrack\) is the conflict coefficient. A higher \(K\) implies stronger disagreement between evidence sources. While Dempster's rule is elegant and widely used, its behavior in high-conflict situations may yield counterintuitive results, motivating a large body of research on conflict management and evidence fusion strategies.

\subsection{Regret theory}

Regret theory (RT) explains choice under uncertainty by embedding the anticipatory emotions of regret and rejoicing. Whereas expected utility theory treats agents as evaluators of absolute payoffs, RT posits that people also compare the realized outcome with what they would have received had they chosen differently. Regret arises when the forgone alternative would have delivered a superior outcome; rejoicing emerges when it would have delivered an inferior one.

\begin{definition} Basic formulation of RT. Let \(A{C}_{1}\) and \(A{C}_{2}\) be two possible actions. If \(A{C}_{1}\) yields outcome \({x}_{1}\) and \(A{C}_{2}\) yields outcome \({x}_{2}\) , the decision maker (DM) does not merely consider the utility values \(u\left( {x}_{1}\right)\) and \(u\left( {x}_{2}\right)\) but also the psychological difference between them. The utility obtained by choosing action \(A{C}_{1}\) and thereby forgoing \(A{C}_{2}\) is \citep{loomes_regret_1982}:  \label{def5} \end{definition}

\begin{equation}\tag{6} \label{eq:6} v = u\left( {x}_{1}\right)  + r\left( {u\left( {x}_{1}\right)  - u\left( {x}_{2}\right) }\right) \end{equation}
where \(u\left( \bullet \right)\) is a monotonic utility function, \(r\left( {\Delta u}\right)\) is an increasing and typically S-shaped regret-rejoice function and is defined as follow:

\begin{definition} Regret-rejoice function. The utility function commonly used in regret theory is \label{def6} \end{definition}

\begin{equation}\tag{7} \label{eq:7} u\left( {x}_{i}\right)  = \frac{1 - {e}^{-\theta {x}_{i}}}{\theta } \end{equation}
where \(\theta  \in  \left( {0,1}\right)\) represents the DM's degree of risk aversion. A widely used parametric representation for regret-rejoice function \citep{bleichrodt_quantitative_2010} is:

\begin{equation}\tag{8} \label{eq:8} r\left( {\Delta u}\right)  = 1 - {e}^{-{\delta \Delta u}} \end{equation}
where \(\delta  \in  \lbrack 0, + \infty )\) denotes the regret-rejoice sensitivity coefficient.

Within this study, RT is repurposed to audit the past behavior of evidence sources. Whenever the fused decision is wrong, each contributor is retrospectively credited with either a regret or a rejoice score: regret if its testimony steered the ensemble away from the truth, rejoice if it pulled the ensemble toward the truth. Aggregating these scores across historical instances quantifies the long-run reliability of every evidence source.

\subsection{Spectral clustering}

Spectral clustering (SC) treats grouping as a graph-partitioning problem: it recovers the intrinsic geometry of data from the eigenvectors of an affinity matrix rather than from distances \citep{ng_spectral_2001}. The algorithm readily separates strongly connected vertex subsets that are only weakly linked to the rest of the graph, giving it a decisive advantage over conventional methods such as k-means when the underlying distribution is high-dimensional \citep{cai_theoretical_2022}.

\begin{definition} Graph construction. Given a dataset \(X = \left\{  {{x}_{1},{x}_{2},\ldots ,{x}_{n}}\right\}\) , each data instance is treated as a vertex in an undirected weighted graph \(G = \langle V,E,W\rangle\) . Edges encode pairwise similarity, and the similarity matrix \(W = \left\lbrack  {w}_{ij}\right\rbrack\) is constructed through a Gaussian kernel: \label{def7} \end{definition}

\begin{equation}\tag{9} \label{eq:9} {w}_{ij} = \exp \left( {-\frac{{\begin{Vmatrix}{x}_{i} - {x}_{j}\end{Vmatrix}}_{2}^{2}}{2{\sigma }^{2}}}\right) \end{equation}
where \(\sigma\) is a scale parameter controlling locality.

\begin{definition} Graph Laplacian Formulation. Given the affinity matrix \(W \in  {\mathbb{R}}^{n \times  n}\) constructed in Definition~\ref{def7}, define the degree matrix \(D\) as a diagonal matrix with entries \({D}_{ii} = \mathop{\sum }\limits_{j}{w}_{ij}\) . The pair \(\left( {D,W}\right)\) induces related Laplacian operators that encode the structural properties of the graph. The symmetric normalized Laplacian is given by \label{def8} \end{definition}

\begin{equation}\tag{10} \label{eq:10} L = I - {D}^{-1/2}W{D}^{-1/2} \end{equation}

\begin{definition} Graph Cut Objective. Given a weighted graph \(G = \left( {V,E,W}\right)\) , clustering can be viewed as partitioning the vertex set into \(k\) disjoint subsets \({g}_{1},{g}_{2},\ldots ,{g}_{k}\) . A desirable partition should minimize the similarity between different subsets while maintaining high within-cluster affinity. To formalize this principle, \citet{shi_normalized_2000} introduced the Normalized Cut (Ncut) criterion, defined for a \(k\) -way partition as \label{def9} \end{definition}

\begin{equation}\tag{11} \label{eq:11} {NCu}{t}_{q} = \mathop{\min }\limits_{{{g}_{1},{g}_{2},\ldots ,{g}_{k} \subseteq  V}}\frac{1}{2}\mathop{\sum }\limits_{{i = 1}}^{k}\frac{{W}_{a}\left( {{g}_{i},\overline{{g}_{i}}}\right) }{\operatorname{vol}\left( {g}_{i}\right) } \end{equation}
where \(W\left( {{g}_{i},\overline{{g}_{i}}}\right)  = \mathop{\sum }\limits_{{u \in  {g}_{i}}}\mathop{\sum }\limits_{{v \in  {g}_{i}}}{w}_{uv}\) denotes the total weight of edges crossing the boundary of \({g}_{i}\) , and \(\operatorname{vol}\left( {g}_{i}\right)  = \mathop{\sum }\limits_{{u \in  {g}_{i}}}{D}_{uu}\) is the degree mass of subset \({g}_{i}\) . The Ncut criterion balances the cut value by the size of each subset, thereby preventing trivial solutions in which a small set is separated from the rest of the graph.

\section{Proposed approach}\label{sec:3_proposed_approach}

\subsection{Overview}

\begin{figure*}[!t]
\centering
\includegraphics[max width=1.0\textwidth]{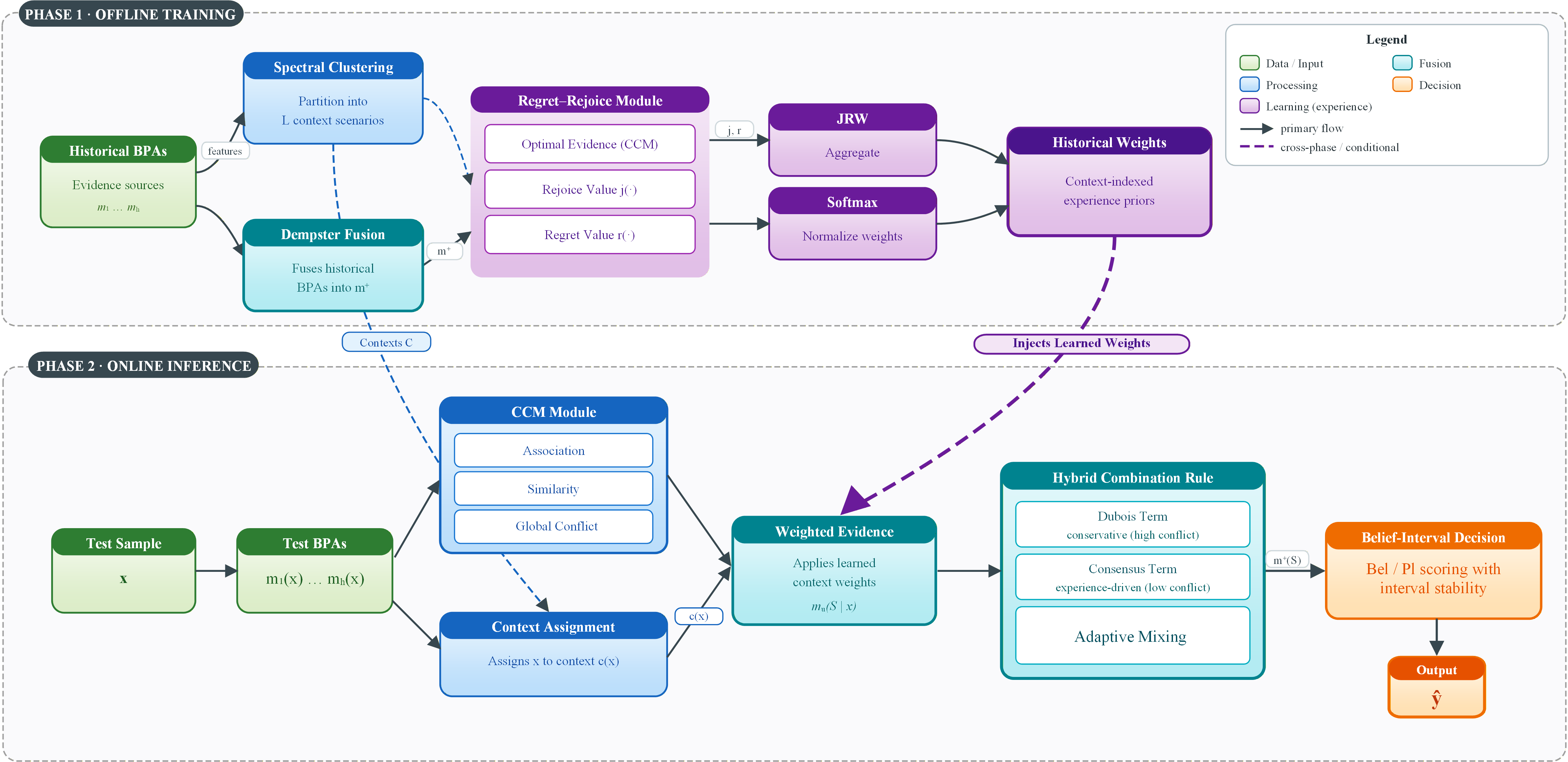}
\caption{Overall architecture of the proposed framework.}
\label{fig:8_255_200_1294_633_0}
\end{figure*}

We propose a DST evidence reasoning framework that combines chaos-conflict measurement with historical-experience weighting for robust multi-source decision making. The framework operates in two phases: offline training and online inference, as illustrated in Figure~\ref{fig:8_255_200_1294_633_0}.

In the offline phase, historical BPAs from multiple evidence sources are processed along two parallel paths. SC partitions the decision space into context scenarios based on Sample feature representations. Meanwhile, Dempster's rule fuses the historical BPAs to identify decision failures. For each failure, the regret-rejoice mechanism evaluates individual evidence sources: the optimal evidence receives a rejoice score proportional to its conflict with the erroneous fusion, while each candidate erroneous source receives a regret score proportional to the distortion it introduces. Aggregated regret and rejoice scores are normalized via softmax to produce context-specific historical experience weights.

In the online phase, a test sample generates BPAs from the same evidence sources. The chaos-conflict measurement computes pairwise association and similarity across all evidence pairs, yielding a global chaos-conflict degree that captures both inter-evidence inconsistency and intra-evidence uncertainty. The test BPAs are then weighted using the context-indexed historical weights retrieved from offline training. A hybrid combination rule adaptively balances a conservative Dubois term with the weighted consensus evidence, controlled by the global conflict level. When conflict is high, the rule emphasizes uncertainty preservation; when conflict is mild, it relies on the consensus term. Finally, a belief-interval decision rule scores each hypothesis by combining its belief lower bound with a stability-weighted plausibility upper bound, outputting the predicted class label.

\subsection{Chaos-conflict measurement}\label{sec:32_chaosconflict_measurement}

In multi-source evidence fusion, conflict reflects discrepancies between evidence sources, whereas uncertainty stems from NFEs. Existing measures treat them independently, yielding incomplete assessments: conflict metrics overlook internal ambiguity, while uncertainty metrics fail to capture cross-evidence inconsistency. To better characterize overall evidential stability, we propose a unified measurement named chaos-conflict measurement (CCM). By jointly evaluating FE set's compatibility and intrinsic uncertainty, CCM offers a more coherent basis for fusion.

\begin{definition} Evidence association measure. Let \({m}_{i}\) and \({m}_{j}\) be two mass functions defined on the same FoD \(\Theta  = \left\{  {{\theta }_{1},{\theta }_{2},\ldots ,{\theta }_{n}}\right\}\) , denote by \(\left\{  {{S}_{1},\ldots ,{S}_{{2}^{n}}}\right\}\) the set of all focal subsets of \(\Theta\) . The pairwise association measure between \({m}_{i}\) and \({m}_{j}\) is defined as  \label{def10} \end{definition}

\begin{equation}\tag{12} \label{eq:12} k\left( {{m}_{i},{m}_{j}}\right)  = \mathop{\sum }\limits_{{i = 1}}^{{2}^{n}}\mathop{\sum }\limits_{{j = 1}}^{{2}^{n}}2\frac{m\left( {S}_{i}\right) m\left( {S}_{j}\right) \left| {{S}_{i} \cap  {S}_{j}}\right| }{\left| {S}_{i}\right| \left| {S}_{j}\right| \left( {\left| {S}_{i}\right|  + \left| {S}_{j}\right| }\right) } \end{equation}

\begin{equation}\tag{13} \label{eq:13} = \mathop{\sum }\limits_{{{S}_{i} \cap  {S}_{j} \neq  \varnothing }}\frac{m\left( {S}_{i}\right) }{\left| {S}_{i}\right| } \cdot  \frac{m\left( {S}_{j}\right) }{\left| {S}_{j}\right| } \cdot  \frac{2\left| {{S}_{i} \cap  {S}_{j}}\right| }{\left| {S}_{i}\right|  + \left| {S}_{j}\right| }
\end{equation}

\begin{corollary} For a collection of mass functions \(\left\{  {{m}_{1},\ldots ,{m}_{J}}\right\}\) defined on the same FoD \(\Theta\) , their global association measure is given by \label{cor1} \end{corollary}

\begin{equation}\tag{13a} \label{eq:13a} k\left( {\underset{j = 1}{\overset{J}{ \oplus  }}{m}_{j}}\right)  = \mathop{\sum }\limits_{{\cap {S}_{i} \neq  \varnothing }}\left( {\mathop{\prod }\limits_{j}\frac{m_j\left( {S}_{i}\right) }{\left| {S}_{i}\right| }}\right) \frac{\left| \mathop{\bigcap }\limits_{i}{S}_{i}\right| }{\mathop{\sum }\limits_{i}\left| {S}_{i}\right| } \end{equation}

Corollary~\ref{cor1} follows immediately from Definition~\ref{def10} by extending the pairwise association to all mass functions on the frame. It aggregates compatible FEs across all evidence sources.

\begin{remark}\label{rem1} The global association measure incorporates both the non-specificity of NFEs and their compatibility structure. In particular, mass assigned to NFEs reflects epistemic uncertainty or ignorance, which weakens the reliability of the corresponding BPA. The formulation therefore down-weights such contributions and adjusts the association strength according to the degree of focal-set overlap. As a result, the measure captures both the uncertainty in NFE allocations and the compatibility across evidence sources, yielding a more faithful characterization of their mutual association.

\end{remark}

\begin{definition} Evidence similarity. For two mass functions \({m}_{i}\) and \({m}_{j}\) defined on the same \(\Theta\) , their similarity is quantified by: \label{def11} \end{definition}

\begin{equation}\tag{14} \label{eq:14} S\left( {{m}_{i},{m}_{j}}\right)  = \frac{k\left( {{m}_{i},{m}_{j}}\right) }{1 - k\left( {{m}_{i},{m}_{j}}\right)  + k\left( {{m}_{i},{m}_{i}}\right) k\left( {{m}_{j},{m}_{j}}\right) } \end{equation}
where \(S\left( \cdot \right)\) denotes the similarity between the two mass functions. Before establishing the main properties of the similarity measure in Definition~\ref{def11}, we first introduce auxiliary lemma that facilitates the subsequent proofs.

\begin{lemma} Cauchy-Schwarz inequality. For any real vectors \(x = {\left( {x}_{1},\ldots ,{x}_{n}\right) }^{T}\) and \(y = {\left( {y}_{1},\ldots ,{y}_{n}\right) }^{T}\) , the following inequality holds: \label{lem1} \end{lemma}

\begin{equation}\tag{15} \label{eq:15} {\left( \mathop{\sum }\limits_{{i = 1}}^{n}{x}_{i}{y}_{i}\right) }^{2} \leq  \left( {\mathop{\sum }\limits_{{i = 1}}^{n}{x}_{i}^{2}}\right) \left( {\mathop{\sum }\limits_{{i = 1}}^{n}{y}_{i}^{2}}\right)
\end{equation}

\noindent with equality if and only if \(x = {\lambda y}\) for some scalar \(\lambda  \in  R\) .

\begin{proof} The result is classical and follows directly from the fact that the quadratic form

\[
\mathop{\sum }\limits_{{i = 1}}^{n}{\left( {x}_{i} - \lambda {y}_{i}\right) }^{2} \geq  0
\]

\noindent holds for all \(\lambda\) .

\end{proof}
\begin{lemma} Positive semi-definiteness of the compatibility matrix. Let \label{lem2} \end{lemma}

\[
{\bar{M}}_{ij} = \frac{2\left| {{S}_{i} \cap  {S}_{j}}\right| }{\left| {S}_{i}\right| \left| {S}_{j}\right| \left( {\left| {S}_{i}\right|  + \left| {S}_{j}\right| }\right) },\;i,j = 1,\ldots ,{2}^{\left| \Theta \right| },
\]

then \(\bar{M}\) is a real symmetric positive semi-definite matrix.

\begin{proof} We begin by fixing an ordering of the power set \({2}^{\Theta },{2}^{\Theta } = \left\{  {{S}_{1},{S}_{2},\ldots ,{S}_{{2}^{\left| \Theta \right| }}}\right\}\) .

It is immediate that \({\bar{M}}_{ij} = {\bar{M}}_{ji}\) , hence \(\bar{M}\) is real and symmetric. Consider any nonzero vector \(z = {\left( {z}_{1},\ldots ,{z}_{{2}^{\left| \Theta \right| }}\right) }^{T}\) , the associated quadratic form is

\[
{z}^{T}\bar{M}z = \mathop{\sum }\limits_{{i = 1}}^{{2}^{\left| \Theta \right| }}\mathop{\sum }\limits_{{j = 1}}^{{2}^{\left| \Theta \right| }}{z}_{i}{z}_{j}\frac{2\left| {{S}_{i} \cap  {S}_{j}}\right| }{\left| {S}_{i}\right| \left| {S}_{j}\right| \left( {\left| {S}_{i}\right|  + \left| {S}_{j}\right| }\right) }.
\]

Because every focal element is compatible with itself, we have

\[
\left| {{S}_{i} \cap  {S}_{i}}\right|  = \left| {S}_{i}\right|  > 0 \Rightarrow  {\bar{M}}_{ii} = \frac{2\left| {S}_{i}\right| }{\left| {S}_{i}\right| \left| {S}_{i}\right| \left( \left| {S}_{i}\right|  + \left| {S}_{i}\right| \right) } = \frac{1}{{\left| {S}_{i}\right| }^{2}} > 0.
\]

Hence each diagonal entry is strictly positive. To show that \(\bar{M}\) is positive semi-definite, we substitute the identity \(\frac{2}{\lvert {S}_{i}\rvert  + \lvert {S}_{j}\rvert } = 2\int_0^\infty {e}^{-\left( {\lvert {S}_{i}\rvert  + \lvert {S}_{j}\rvert }\right) t}\,dt\) into the quadratic form, which yields

\[
{z}^{T}\bar{M}z = 2\int_0^\infty \sum_{\theta \in \Theta }{\left( \sum_{i \ni \theta }{z}_{i}\frac{{e}^{-\lvert {S}_{i}\rvert t}}{\lvert {S}_{i}\rvert }\right) }^{2}\,dt \geq  0
\]

for every vector \(z\), because the integrand is a sum of squares. Hence \({z}^{T}\bar{M}z \geq 0\) for all \(z\), which proves that \(\bar{M}\) is positive semi-definite.

\end{proof}
We adopted the set of attributes that a conflict measure ought to satisfy as stipulated by \citet{destercke_toward_2013} and supply the corresponding proof below.
\begin{property} Boundedness, \(0 \leq  S\left( {{m}_{i},{m}_{j}}\right)  \leq  1\) . \label{pro1} \end{property}

\begin{proof} Since \(\bar{M}\) is positive semi-definite, there exists a lower-triangular matrix \(L\) such that \(\bar{M} = L{L}^{\top }\) . Define \(\mathbf{x} = \mathbf{v}L,\mathbf{y} = \mathbf{b}L\) . Then \(k\left( {{m}_{i},{m}_{j}}\right)  = 2\mathbf{v}L{L}^{\top }{\mathbf{b}}^{\top } = 2{\mathbf{{xy}}}^{\top },k\left( {{m}_{i},{m}_{i}}\right)  = 2{\mathbf{{xx}}}^{\top } \; k\left( {{m}_{j},{m}_{j}}\right)  = 2{\mathbf{{yy}}}^{\top }\) . Set \(a = {\mathbf{{xy}}}^{\top },b = {\mathbf{{xx}}}^{\top },c = {\mathbf{{yy}}}^{\top }\) . Then

\[
S\left( {{m}_{i},{m}_{j}}\right)  = \frac{2a}{1 - {2a} + {4bc}}.
\]

By construction, all entries of \(\bar{M}\) and of the mass vectors are non-negative, hence \(a,b,c \geq  0\) . Moreover \(0 \leq  k\left( {{m}_{i},{m}_{j}}\right)  \leq  1\) and \(k\left( {{m}_{i},{m}_{j}}\right)  = {2a}\) imply \(a \in  \left\lbrack  {0,1/2}\right\rbrack\) . Thus the denominator satisfies \(1 - {2a} + {4bc} \geq  1 - {2a} \geq  0\) , and the numerator \({2a} \geq  0\) ; therefore \(S\left( {{m}_{i},{m}_{j}}\right)  \geq  0\) .

By Lemma~\ref{lem1}, \({a}^{2} = {\left( {\mathbf{{xy}}}^{\top }\right) }^{2} \leq  \left( {\mathbf{{xx}}}^{\top }\right) \left( {\mathbf{{yy}}}^{\top }\right)  = {bc}\) . Let \(u = {bc} \geq  0\) and \(0 \leq  a \leq  \sqrt{u}\) . To show \(S\left( {{m}_{i},{m}_{j}}\right)  \leq  1\) , it suffices to prove \({2a} \leq  1 - {2a} + {4bc} \Leftrightarrow  0 \leq  u - \sqrt{u} + 1/4\) . Consider the function \(f\left( u\right)  = u - \sqrt{u} + 1/4\) , Its derivative is

\[
{f}^{\prime }\left( u\right)  = 1 - \frac{1}{2\sqrt{u}},u \in  (0,1\rbrack .
\]

The equation \({f}^{\prime }\left( u\right)  = 0\) has the unique solution \({u}^{ * } = 1/4\) . Since \(f\left( {1/4}\right)  = 0\) , by continuity we obtain \(f\left( u\right)  \geq  0,\;\forall u \in  \left\lbrack  {0,1}\right\rbrack\) . Therefore \(S\left( {{m}_{i},{m}_{j}}\right)  \leq  1\) .

\end{proof}
\begin{property} Symmetry, \(S\left( {{m}_{i},{m}_{j}}\right)  = S\left( {{m}_{j},{m}_{i}}\right)\) . \label{pro2} \end{property}

\begin{proof}
\[S\left( {{m}_{i},{m}_{j}}\right)  = \frac{k\left( {{m}_{i},{m}_{j}}\right) }{1 - k\left( {{m}_{i},{m}_{j}}\right)  + k\left( {{m}_{i},{m}_{i}}\right)  \cdot  k\left( {{m}_{j},{m}_{j}}\right) }
\]

\begin{multline*}
= \mathop{\sum }\limits_{{i = 1}}^{{2}^{n}}\mathop{\sum }\limits_{{j = 1}}^{{2}^{n}}2\frac{m\left( {S}_{i}\right) m\left( {S}_{j}\right) \left| {{S}_{i} \cap  {S}_{j}}\right| }{\left| {S}_{i}\right| \left| {S}_{j}\right| \left( {\left| {S}_{i}\right|  + \left| {S}_{j}\right| }\right) } \\
= \mathop{\sum }\limits_{{i = 1}}^{{2}^{n}}\mathop{\sum }\limits_{{j = 1}}^{{2}^{n}}2\frac{m\left( {S}_{j}\right) m\left( {S}_{i}\right) \left| {{S}_{j} \cap  {S}_{i}}\right| }{\left| {S}_{j}\right| \left| {S}_{i}\right| \left( {\left| {S}_{j}\right|  + \left| {S}_{i}\right| }\right) } = S\left( {{m}_{j},{m}_{i}}\right)
\end{multline*}

Thus, the similarity measure is symmetric.

\end{proof}

\begin{property}\label{pro3} Monotonicity. Let \({m}_{1}\) and \({m}_{2}\) be two mass functions defined on the same FoD \(\Theta\) . Without loss of generality, assume \(k\left( {{m}_{2},{m}_{2}}\right)  \geq  k\left( {{m}_{1},{m}_{1}}\right)\) . For \(t \in  \left\lbrack  {0,1}\right\rbrack\) , consider the convex combination \({m}_{3}\left( t\right)  = \left( {1 - t}\right) {m}_{2} + t{m}_{1}\) , define \(\zeta \left( t\right)  \mathrel{\text{ := }} S\left( {{m}_{1},{m}_{3}\left( t\right) }\right)\) . Then \(\zeta \left( t\right)\) is non-decreasing on \(\left\lbrack  {0,1}\right\rbrack\) . Moreover, if \({m}_{1} \neq  {m}_{2}\) , then \(\zeta \left( t\right)\) is strictly increasing on \(\left( {0,1}\right)\) . \end{property}

\begin{proof} By Property~\ref{pro1}, we have

\[
\frac{\partial S}{\partial a} = \frac{2 + {8bc}}{{\left( 1 - 2a + 4bc\right) }^{2}} > 0.
\]

Thus, \(S\left( {{m}_{i},{m}_{j}}\right)\) is a strictly increasing transform of \(k\left( {{m}_{i},{m}_{j}}\right)\) . Meanwhile,

\[
k\left( {{m}_{1},{m}_{3}\left( t\right) }\right)  = {2a}\left( t\right)  \propto  {xy}{\left( t\right) }^{\top } = x{\left( \left( 1 - t\right) y + tx\right) }^{\top }
\]

\[
= \left( {1 - t}\right) x{y}^{\top } + {tx}{x}^{\top } = \left( {1 - t}\right) a + {tb} = {\left\langle  {m}_{1},\left( 1 - t\right) {m}_{2} + t{m}_{1}\right\rangle  }_{D}.
\]

Differentiating gives \({a}^{\prime }\left( t\right)  = b - a\) . Because \(a \leq  \sqrt{bc}\) and \(b \geq  c,{a}^{\prime }\left( t\right)  > 0,\;{m}_{i} \neq  {m}_{j}\) . The composition \(\;S \circ  a\;\) is strictly increasing. Consequently, \({t}_{1} < {t}_{2} \Rightarrow  S\left( {{m}_{1},{m}_{3}\left( {t}_{1}\right) }\right)  < S\left( {{m}_{1},{m}_{3}\left( {t}_{2}\right) }\right) .\)

\end{proof}
\begin{property} Extreme Consistency. (1) If and only if \({m}_{i} = {m}_{j}\) and the BPAs of \({m}_{i}\) and \({m}_{j}\) are all concentrated on the same SFE, \(S\left( {{m}_{i},{m}_{j}}\right)  = 1\) ;(2)\(S\left( {{m}_{i},{m}_{j}}\right)  = 0 \Leftrightarrow  \left( {\bigcup {S}_{i}}\right)  \cap  \left( {\bigcup {S}_{j}}\right)  = \varnothing .\) \label{pro4} \end{property}

\begin{proof} For (1), by Property~\ref{pro1}, we have \(a = 1/4 + {bc}\) and \({a}^{2} \leq  {bc}\) . When \({u}^{ * } = 1/4\) , we have \(a = b = c = 1/2\) , then \(x = y\) and \(k\left( {{m}_{i},{m}_{j}}\right)  = k\left( {{m}_{i},{m}_{i}}\right)  = k\left( {{m}_{j},{m}_{j}}\right)  = 1\) , condition \({m}_{i} = {m}_{j}\) is satisfied. Due to Eq.~\ref{eq:2}, let the unique FE be \(A\) , then their masses are \({m}_{i}\left( A\right)  = {m}_{j}\left( A\right)  = 1\) . If \(A\) is a NFE, \({\left( 1/\left| A\right| \right) }^{2} < 1\) , which contradicts \(k\left( {{m}_{i},{m}_{j}}\right)  = 1\) . Thus \(A\) must be a SFE. For (2), by the Property \(1,S\left( {{m}_{i},{m}_{j}}\right)  = 0 \Leftrightarrow  a = 0\) . Due to the construction, we have \(a \propto  \mathop{\sum }\limits_{{{S}_{i} \cap  {S}_{j} \neq  \varnothing }}{m}_{i}\left( {S}_{i}\right) {m}_{j}\left( {S}_{j}\right) \left| {{S}_{i} \cap  {S}_{j}}\right| /\left( {\left| {S}_{i}\right|  + \left| {S}_{j}\right| }\right)\) and all terms in the summation are nonnegative. Therefore \(a\) can be zero only if \(\left| {{S}_{i} \cap  {S}_{j}}\right|  = 0\) , which is equivalent to

\[
\left( {\cup {S}_{i}}\right)  \cap  \left( {\cup {S}_{j}}\right)  = \varnothing \text{ . }
\]

\end{proof}
\begin{property} Refinement Insensitivity. When the FoD is refined from \(\Theta\) to \({\Theta }^{\prime },S\left( {{m}_{i},{m}_{j}}\right)\) remains unchanged. \label{pro5} \end{property}

\begin{proof} A refinement of the frame preserves the structure of both mass functions. Let \(S \subseteq  {\Theta }^{\prime }\) denote such a refined FE. Under refinement, the masses satisfy \({m}_{i}\left( S\right)  = {m}_{j}\left( S\right)  = 0\) , and these terms appear symmetrically in the computation of \(k\left( {{m}_{i},{m}_{j}}\right)\) , contributing additively as zeros. Removing these null contributions does not affect any terms involved in the definition of the similarity measure. Consequently, the value of \(S\left( {{m}_{i},{m}_{j}}\right)\) is invariant under refinement of the frame.

\end{proof}
To further illustrate the proposed similarity measure, we consider two numerical examples.

\begin{figure}[H]
\centering
\includegraphics[max width=\columnwidth]{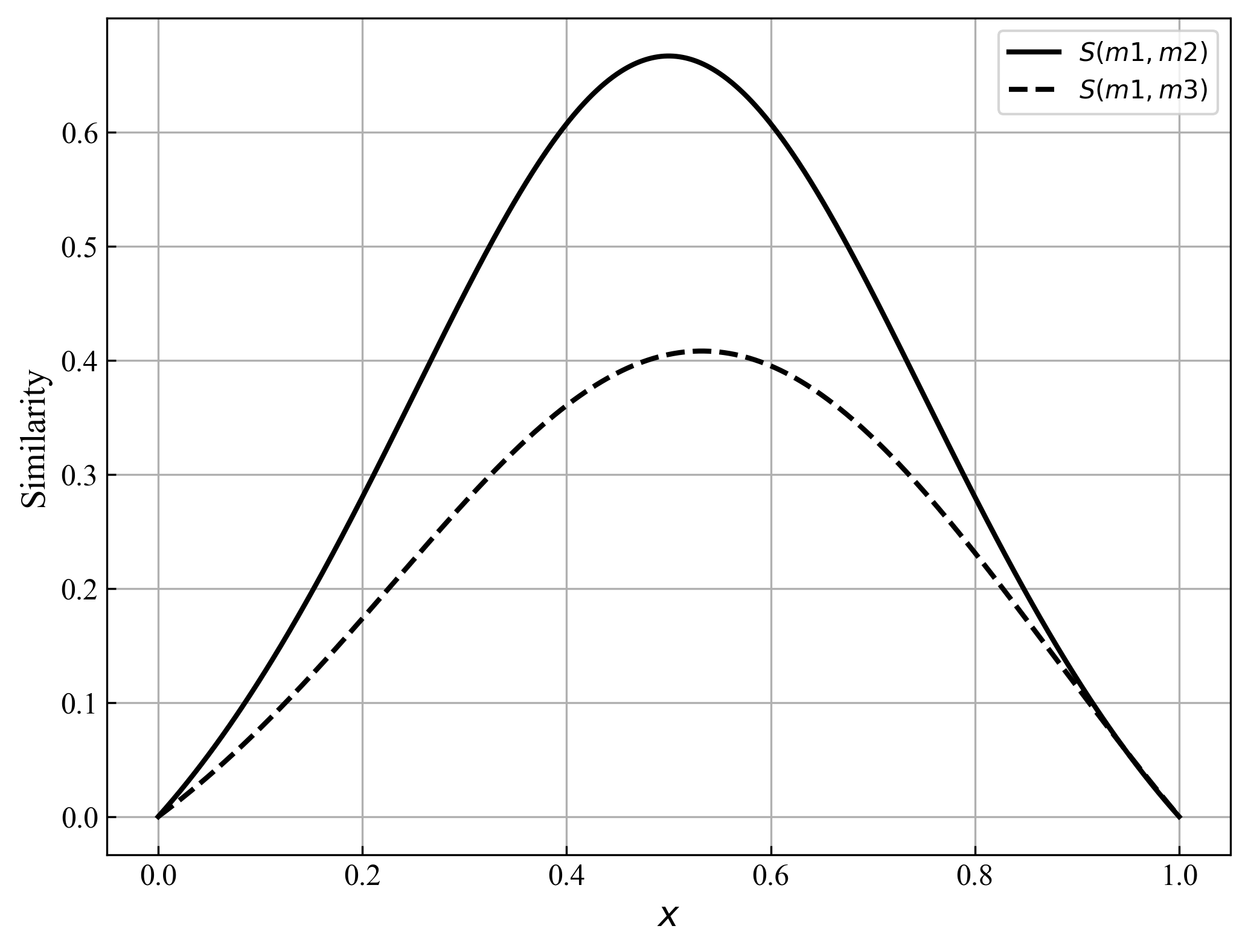}
\caption{Similarity for Example 1}
\label{fig:14_270_215_1263_952_0}
\end{figure}

\begin{example} Three BPA functions \({m}_{1},{m}_{2}\) and \({m}_{3}\) are defined on the same FoD \(\Theta\) and constructed as follows:

\[
{m}_{1}\left( {\theta }_{1}\right)  = x,{m}_{1}\left( {\theta }_{2}\right)  = 1 - x
\]

\[
{m}_{2}\left( {\theta }_{1}\right)  = 1 - x,{m}_{2}\left( {\theta }_{2}\right)  = x
\]

\[
{m}_{3}\left( {\theta }_{1}\right)  = 1 - x,{m}_{3}\left( {{\theta }_{1},{\theta }_{2}}\right)  = x
\]

\noindent where \(x \in  \left\lbrack  {0,1}\right\rbrack\) . Using Eqs.~\ref{eq:12} and~\ref{eq:14}, the similarity values are computed and the results are shown in Figure~\ref{fig:14_270_215_1263_952_0}. Several observations can be made:

(1) When \(x = 0\) or \(x = 1\) , the BPAs become completely conflicting, the similarity correctly returns \(S\left( \cdot \right)  = 0\) ;

(2) When \(x = 0.5\), \({m}_{1}\) and \({m}_{2}\) become identical. However, the resulting similarity does not reach 1. This is because \(S\left( \cdot \right)\) not only considers the closeness between the two BPAs, but also incorporates their intrinsic uncertainty (nonspecificity). At this moment, both BPAs carry the maximum amount of ambiguity and contribute no effective decision-making information. This observation is consistent with Property~\ref{pro4}, which establishes that similarity reaches 1 only under defined conditions. And moreover:

(3) We observe that \(S\left( {{m}_{1},{m}_{3}}\right)  \leq  S\left( {{m}_{1},{m}_{2}}\right)\) . Compared with \({m}_{2},{m}_{3}\) assigns part of its mass to NFE. This allocation increases the level of nonspecificity, resulting in greater intrinsic uncertainty. Consequently, even when the BPA values appear similar, the uncertainty prevents the similarity from reaching a higher value.

\end{example}

\begin{figure}[H]
\centering
\includegraphics[max width=\columnwidth]{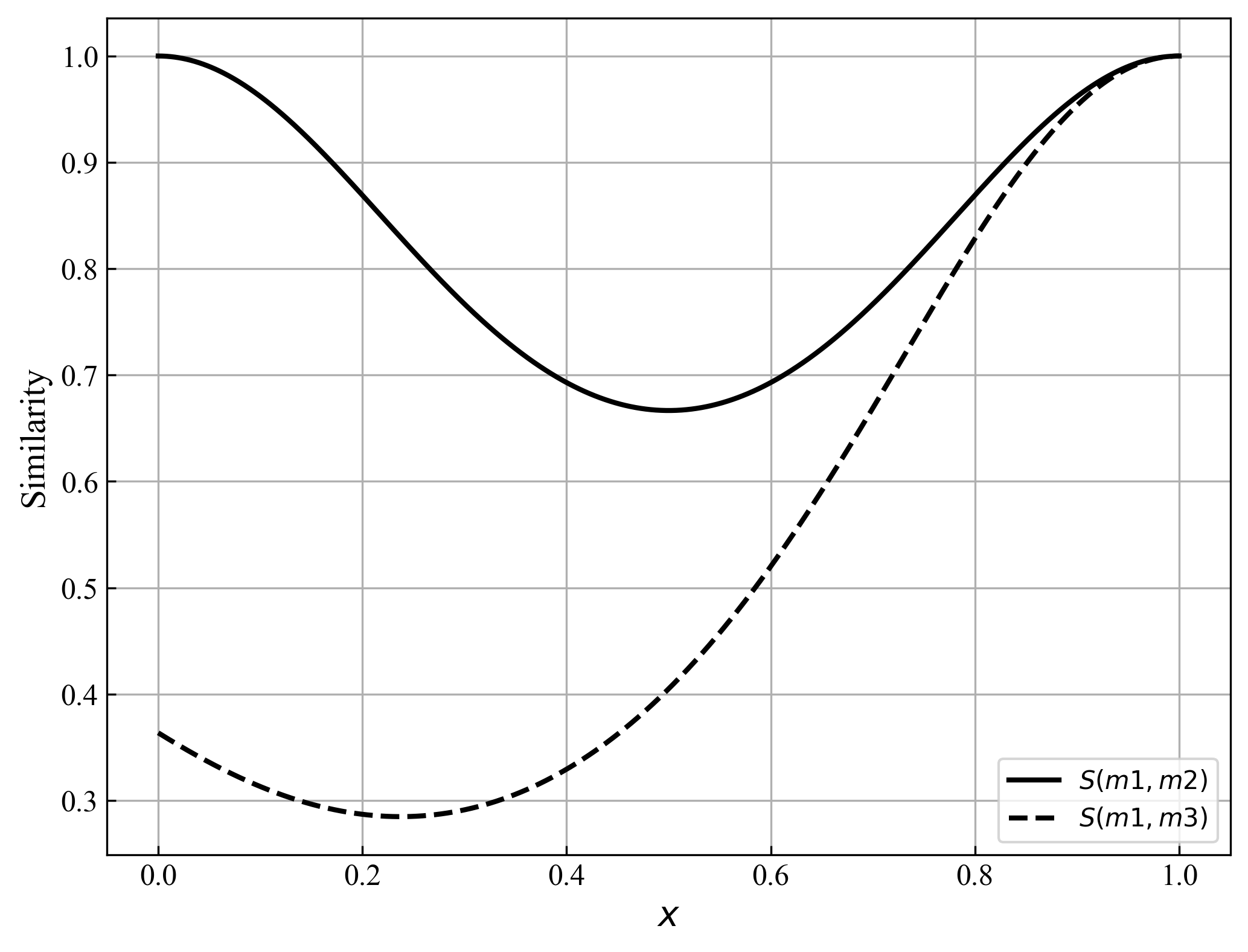}
\caption{Similarity for Example 2}
\label{fig:15_270_215_1264_953_0}
\end{figure}

\begin{example}

\[
{m}_{1}\left( {\theta }_{1}\right)  = x,{m}_{1}\left( {\theta }_{2}\right)  = 1 - x
\]

\[
{m}_{2}\left( {\theta }_{1}\right)  = x,{m}_{2}\left( {\theta }_{2}\right)  = 1 - x
\]

\[
{m}_{3}\left( {\theta }_{1}\right)  = x,{m}_{3}\left( {{\theta }_{1},{\theta }_{2}}\right)  = 1 - x
\]

\noindent where \(x \in  \left\lbrack  {0,1}\right\rbrack\) and the similarity results are shown in Figure~\ref{fig:15_270_215_1264_953_0}. From Figure~\ref{fig:15_270_215_1264_953_0}, it can be observed that even when two evidences follow identical BPA assignment, their similarity values may still differ significantly. Specifically, when the BPA becomes more dispersed, the associated uncertainty increases accordingly. In such cases, it becomes difficult to reliably assess the similarity between two evidences, as higher uncertainty weakens the confidence in their consistency. Furthermore, when BPA values are transferred from SFEs to their supersets, as in the case of \({m}_{3}\) , additional uncertainty is introduced. Since a NFE contains multiple propositions, the belief mass assigned to it cannot be fully committed to any single hypothesis. As a result, the evidence provides less information for decision-making, leading to a lower similarity.

\end{example}

In summary, the proposed similarity measure quantifies the agreement between evidences by jointly comparing their BPA while fully accounting for their intrinsic uncertainty. On this basis, we introduce the CCM of evidence as follows:

\begin{definition} CCM. Let \({m}_{i}\) and \({m}_{j}\) be two arbitrary mass functions defined on the same FoD \(\Theta\) and the CCM is defined as \label{def12} \end{definition}

\begin{multline}\tag{15a} \label{eq:ccm} \widehat{K}\left( {{m}_{i},{m}_{j}}\right)  = 1 - S\left( {{m}_{i},{m}_{j}}\right) \\
 = \frac{1 - {2k}\left( {{m}_{i},{m}_{j}}\right)  + k\left( {{m}_{i},{m}_{i}}\right)  \cdot  k\left( {{m}_{j},{m}_{j}}\right) }{1 - k\left( {{m}_{i},{m}_{j}}\right)  + k\left( {{m}_{i},{m}_{i}}\right)  \cdot  k\left( {{m}_{j},{m}_{j}}\right) } \end{multline}

It is straightforward to verify that the CCM inherits the same five desirable properties as the similarity measure.

\begin{corollary} Global chaos-conflict degree (GCCD). Let \(\left\{  {{m}_{1},{m}_{2},\ldots ,{m}_{n}}\right\}\) be a set of mass functions defined on the same FoD. The GCCD of the evidence set is defined as \label{cor2} \end{corollary}

\begin{equation}\tag{16} \label{eq:16} \widehat{K} = \frac{1}{n\left( {n - 1}\right) /2}\mathop{\sum }\limits_{{i < j}}\widehat{K}\left( {{m}_{i},{m}_{j}}\right) . \end{equation}

\begin{corollary} Relative stability reliability. Given a set of mass functions \(\left\{  {{m}_{1},{m}_{2},\ldots ,{m}_{n}}\right\}\) defined on the same FoD, the relative stability reliability of an individual evidence source \({m}_{i}\) is defined as \label{cor3} \end{corollary}

\begin{equation}\tag{17} \label{eq:17} {SR}\left( {m}_{i}\right)  = \frac{\mathop{\sum }\limits_{{j \neq  i}}S\left( {{m}_{i},{m}_{j}}\right) }{1 - \widehat{K}}. \end{equation}

While the above definitions and corollaries enable instantaneous evidence assessment of conflict and stability, they remain limited to the current decision context. In practical applications where evidence sources are repeatedly reused under heterogeneous conditions, exploiting historical experience becomes essential for achieving robust and adaptive evidence fusion, which is the focus of the next section.

\subsection{Evidence weighting driven by historical experience}\label{sec:33}

Existing strategies predominantly assess evidential inconsistency based on the current hypothesis space and the configuration of BPAs. Such approaches, however, implicitly assume that all evidence sources are equally reliable across different decision contexts \citep{Qiu_2025}, thereby overlooking the informative value embedded in their long-term historical performance. In real-world applications, evidence sources are rarely one-shot contributors. The generation of BPAs is often constrained by complex environmental factors, and the reliability of an evidence source typically exhibits within specific scene. Relying exclusively on instantaneous conflict evaluation may therefore lead to biased reliability assessment and suboptimal evidence weighting \citep{Zhang_2025,Guo_2026}. A representative example can be found in complex diagnostic scenarios, where different evidence sources exhibit markedly different detection capabilities across specific conditions \citep{Lu_2026}. In such cases, certain sources may systematically outperform others in particular contexts, despite occasionally producing conflicting assessments. Treating these discrepancies solely as instantaneous conflict may overlook the historically validated reliability of informative evidence sources and lead to their unjustified suppression during fusion \citep{Wang_2021}.

Motivated by these observations, we introduce a historical experience driven evidence weighting mechanism that complements the CCM proposed above. Rather than relying exclusively on instantaneous conflict evaluation, the proposed approach leverages historical decision outcomes to learn context-dependent evidence weights, which are subsequently integrated into a hybrid combination rule that adaptively balances conservative uncertainty preservation with weighted consensus.

Specifically, heterogeneous decision contexts are first identified via clustering, after which the historical behavior of each evidence source is evaluated through regret and rejoice measures derived from regret theory. These measures quantify the counterfactual contribution of individual evidence sources when fusion outcomes deviate from ground truth, enabling continuous credit assignment that rewards evidence sources aligned with successful outcomes and penalizes those contributing to fusion errors. The accumulated historical experience is then translated into adaptive evidence weights, which are subsequently incorporated into the fusion process.

By explicitly integrating long-term evidence behavior with instantaneous conflict assessment, our approach avoids indiscriminate suppression of conflicting evidences and yields a more reliable fusion strategy.

RT is employed to analyze the potential regret effects that may arise when decision outcomes deviate from expectations. This theory not only focuses on the absolute gains of decisions but also emphasizes the relative gains compared to other potential choices. Introducing regret theory into the calculation of historical experience for evidence synthesis provides a quantitative and systematic method for evaluating the applicability of different pieces of evidence in historical contexts, thereby aiding in the optimization of future evidence selection and integration strategies. Among these, the calculation method for regret-rejoicing values, which assesses the applicability of evidence, will serve as the core research focus of this section. Next, we will introduce the regret-rejoicing measurement of evidence bodies during the evidence synthesis process. Let \(\Theta  = \left\{  {{\theta }_{1},\ldots ,{\theta }_{n}}\right\}\) denote the FoD and \(M = \left\{  {{m}_{1},{m}_{2},\ldots ,{m}_{h}}\right\}\) be the set of available evidence sources defined on \({2}^{\Theta }\) .

\begin{definition} Optimal evidence. For a given sample whose ground-truth hypothesis is known, the optimal evidence is defined as the evidence source whose BPA is most consistent with the truth according to the similarity measure defined in Eq.~\ref{eq:14}: \label{def13} \end{definition}

\begin{equation}\tag{18} \label{eq:18} {m}_{\text{ best }} = \arg \mathop{\min }\limits_{{{m}_{i} \in  M}}\widehat{K}\left( {{m}_{i},{p}_{\text{ target }}}\right) \end{equation}

Here, \({p}_{\text{ target }}\) is a one-hot vector of dimension \({2}^{n}\) .

\begin{definition} Error-removal combination. Given the evidence set \(M\) , the initial fused result \({m}_{ \oplus  }\) is first obtained using Dempster's rule of combination as defined in Eq.~\ref{eq:5}. If the decision derived from \({m}_{ \oplus  }\) is inconsistent with the truth, a decision error is considered to have occurred. In such a case, the evidence set is assumed to contain at least one erroneous evidence source, whose mass function fails to assign its maximum support to the true hypothesis. \label{def14} \end{definition}

Let \({m}_{e} \in  M\) denote a candidate erroneous evidence source. By removing \({m}_{e}\) from the \(M\) and recombining the remaining evidence, the error-removal combination result \({m}_{ \oplus  }^{-e}\) is defined as

\begin{equation}\tag{19} \label{eq:19} \left\{  {\begin{array}{l} {m}_{ \oplus  }^{-e}\left( S\right)  = \frac{\mathop{\sum }\limits_{{\cap {S}_{j} = S,{S}_{j} \subseteq  {2}^{\Theta }}}\mathop{\prod }\limits_{{i \neq  e}}{m}_{i}\left( {S}_{j}\right) }{1 - K} \\  K = \mathop{\sum }\limits_{{\bigcap {S}_{j} = \varnothing ,{S}_{j} \subseteq  {2}^{\Theta }}}\mathop{\prod }\limits_{{i \neq  e}}{m}_{i}\left( {S}_{j}\right)  \end{array}.}\right. \end{equation}

\begin{definition} Rejoice value of the optimal evidence. Based on the CCM and the formulation of RT, the rejoice value associated with the \({m}_{\text{ best }}\) is defined as \label{def15} \end{definition}

\begin{equation}\tag{20} \label{eq:20} j\left( {m}_{\text{ best }}\right)  = 1 - \exp \left( {-\eta \widehat{K}\left( {{m}_{\text{ best }},{m}_{ \oplus  }}\right) }\right) , \end{equation}
where \(\eta  \in  (0,1\rbrack\) is a sensitivity parameter and controls the responsiveness of the rejoice value to variations in conflict intensity.

\begin{remark}\label{rem2} The underlying intuition of Definition~\ref{def15} is as follows. When the fused evidence leads to an incorrect decision, it indicates that the collective judgment derived from evidence aggregation is unreliable. In such cases, the decision supported by the optimal evidence may provide a more faithful reflection of ground truth. Consequently, a larger conflict between \({m}_{\text{ best }}\) and the erroneous fusion result \({m}_{ \oplus  }\) implies that \({m}_{\text{ best }}\) is closer to the true outcome, and should therefore be assigned a higher rejoice value. \end{remark}

\begin{definition} Regret value of erroneous evidence. Analogous to Definition~\ref{def15}, the regret value associated with an erroneous evidence source \({m}_{e}\) is defined as \label{def16} \end{definition}

\begin{equation}\tag{21} \label{eq:21} r\left( {m}_{e}\right)  = 1 - \exp \left( {-\gamma \widehat{K}\left( {{m}_{ \oplus  }^{-e},{m}_{ \oplus  }}\right) }\right) , \end{equation}
where \(\gamma  \in  (0,1\rbrack\) is a sensitivity parameter analogous to \(\eta\) , which governs the responsiveness of the regret value to variations in conflict intensity.

\begin{remark}\label{rem3} The intuitive interpretation of Definition~\ref{def16} is that the discrepancy between the \({m}_{ \oplus  }\) and the \({m}_{ \oplus  }^{-e}\) reflects the extent to which the erroneous evidence distorts the collective decision. A larger conflict between these two fusion outcomes indicates a stronger negative impact of \({m}_{e}\) on the fusion process, and consequently corresponds to a higher regret value. \end{remark}

In practical decision systems, the reliability of an evidence source is rarely invariant across all operating conditions. Instead, evidence behavior typically exhibits strong context dependency: a source that is highly informative in one scenario may become less reliable in another due to changes in acquisition conditions, latent nuisance factors, or feature distribution shifts \citep{Huang_2025,Zhao_2025}. To reflect such heterogeneity, we partition the historical samples into a set of decision contexts (clusters), and assign each sample to a context. Within each context, we then aggregate regret and rejoice values induced by decision failures, thereby obtaining context-conditioned historical experience for each evidence.

\begin{definition} Historical experience-based evidence weight. Let \(\mathcal{H} = {\left\{  \left( {x}_{t},{y}_{t}\right) \right\}  }_{t = 1}^{N}\) be the historical set and \(\mathcal{C} = \left\{  {{c}_{1},\ldots ,{c}_{L}}\right\}\) the context partition induced by clustering algorithm. \(c\left( {x}_{t}\right)  \in  \mathcal{C}\) denotes the cluster membership of \({x}_{t}\) . For an evidence source \({m}_{i}\) and a target context \(c \in  \mathcal{C}\) , define the failure indicator \label{def17} \end{definition}

\begin{equation}\tag{22} \label{eq:22} {\delta }_{t} = \mathbb{I}\left\lbrack  {\arg \mathop{\max }\limits_{{{\theta }_{k} \in  \Theta }}{Be}{l}_{{m}_{ \oplus  }\left( {\cdot  \mid  {x}_{t}}\right) }\left( {\theta }_{k}\right)  \neq  {y}_{t}}\right\rbrack \end{equation}
where \(\mathbb{I}\left( \cdot \right)\) denotes the indicator function and \({\delta }_{t} = 1\) indicates that the final fusion decision at sample \({x}_{t}\) is incorrect. Let \({j}_{t}\left( {m}_{i}\right)\) and \({r}_{t}\left( {m}_{i}\right)\) denote the rejoice and regret values of evidence source \({m}_{i}\) computed according to Eqs.~\ref{eq:20} and~\ref{eq:21}, the weight score of \({m}_{i}\) under context \(c\) is defined as:

\begin{equation}\tag{23} \label{eq:23} {JR}{W}_{i}^{c} = \mathop{\sum }\limits_{{t = 1}}^{N}{\delta }_{t}\mathbb{I}\left\lbrack  {c\left( {x}_{t}\right)  = c}\right\rbrack {j}_{t}\left( {m}_{i}\right)  - \mathop{\sum }\limits_{{t = 1}}^{N}{\delta }_{t}\mathbb{I}\left\lbrack  {c\left( {x}_{t}\right)  = c}\right\rbrack {r}_{t}\left( {m}_{i}\right) . \end{equation}

Then, softmax normalization

\begin{equation}\tag{24} \label{eq:24} \widetilde{{JR}{W}_i^{c}} = \frac{\exp \left( {{JR}{W}_{i}^{c}}\right) }{\mathop{\sum }\limits_{{j = 1}}^{M}\exp \left( {{JR}{W}_{j}^{c}}\right) } \end{equation}

\noindent is implemented to guarantee the final weight \(\widetilde{{JR}{W}_i^{c}} \geq  0\) and \(\mathop{\sum }\limits_{i}\widetilde{{JR}{W}_i^{c}} = 1\) .

\subsection{A robust fusion-and-decision rule}\label{sec:34}

Building upon the historical experience-driven evidence weighting mechanism developed in Section~\ref{sec:33}, we now construct a robust fusion rule that integrates long-term source reliability with instantaneous evidential conflict. Specifically, Section~\ref{sec:33} yields context-dependent reliability profiles learned from historical decision outcomes, which enable the construction of a weighted evidence representation reflecting the heterogeneous performance of different evidence sources. While such historical calibration mitigates the risk of indiscriminately suppressing informative evidence, robust fusion further requires an explicit treatment of the conflict structure exhibited by the current evidence set. To this end, Section~\ref{sec:32_chaosconflict_measurement} introduced the CCM, which characterizes not only the degree of evidential inconsistency but also the uncertainty embedded in evidence formation.

Motivated by these observations, this section proposes a new robust evidence fusion rule following a principled two-level design. First, instead of aggregating evidence via the conventional simple averaging strategy in \citet{murphy_combining_2000}'s method, we employ reliability discounting guided by historical experience to obtain a calibrated weighted evidence representation. This calibrated evidence is subsequently used as the consensus component in the fusion process. Second, the CCM is computed to quantify the instantaneous inconsistency more faithfully. Finally, a conflict-adaptive hybrid combination is performed, which smoothly balances informative consensus and uncertainty according to the estimated conflict level. In this way, the proposed rule achieves improved robustness under heterogeneous contexts. The following is the historical experience weighted evidence we obtained:

\begin{equation}\tag{25} \label{eq:25} {m}_{w}\left( {S \mid  x}\right)  = \mathop{\sum }\limits_{i}\widetilde{{JR}{W}_i^{c(x)}} \cdot  {m}_{i}\left( {S \mid  x}\right) \end{equation}
where \(x\) denotes the target sample for which evidence fusion is performed, and the notation \({m}_{i}\left( {S \mid  x}\right)\) indicates that the BPA provided by the \(\underline{i}\) -th evidence source. The corresponding historical experience weight \(\widetilde{{JR}{W}_i^{c(x)}}\) is therefore selected, ensuring that the contribution of each evidence is modulated according to its long-term performance under similar conditions. Consequently, the weighted evidence \({m}_{w}\left( {S \mid  x}\right)\) represents a context-aware aggregation of the original BPAs associated with the sample \(x\) . After obtaining weighted evidence based on historical experience, we give a new evidence fusion rule defined as follows.

\begin{definition} Hybrid combination rule. To construct belief intervals and to allocate the aggregated mass more informatively, we define a hybrid combination rule as follows: \label{def18} \end{definition}

\begin{multline}\tag{26} \label{eq:26} {m}^{ \oplus  }\left( S\right)  = \left( {1 - {e}^{-\widehat{K}}}\right) \Biggl(\mathop{\sum }\limits_{{{A}_{1}\bigcap \cdots \bigcap {A}_{M} = S \neq  \varnothing }}\mathop{\prod }\limits_{{i = 1}}^{M}{m}_{i}\left( {A}_{i}\right) \\
  + \mathop{\sum }\limits_{{\substack{{A}_{1}\bigcap \cdots \bigcap {A}_{M} = \varnothing \\ {A}_{1}\bigcup \cdots \bigcup {A}_{M} = S}}}\mathop{\prod }\limits_{{i = 1}}^{M}{m}_{i}\left( {A}_{i}\right) \Biggr) + {e}^{-\widehat{K}}{m}_{w}\left( S\right) \end{multline}

Eq.~\ref{eq:26} consists of two complementary terms. The first term is a refined Dubois combination rule \citet{liu_synergy_2023}, which aggregates both conjunctive contributions and union-based allocations so as to explicitly preserve uncertainty under severe disagreement. The second term \({m}_{w}\left( S\right)\) represents the consensus evidence obtained by weighted averaging, which is beneficial when the evidence sources are largely consistent. The mixing factor \({e}^{-\widehat{K}} \in  (0,1\rbrack\) adaptively controls the trade-off: when the overall conflict is high, \({e}^{-\widehat{K}}\) becomes small and the rule places greater emphasis on the conservative term, thereby pushing more mass toward NFEs and retaining uncertainty for subsequent interval construction; when the conflict is mild, the rule increasingly relies on the consensus term, yielding a sharper and more decisive fused belief assignment.

Having obtained the final fused mass function through the proposed Definition~\ref{def18}, a principled decision rule is still required to output a unique hypothesis. This step is nontrivial in the presence of composite FEs, since the fused evidence may preserve a nonnegligible amount of epistemic uncertainty. To this end, we adopt a belief interval-based decision strategy, which directly exploits the confidence bounds induced. The resulting score provides a measure for each singleton hypothesis and enables a deterministic decision.

\begin{definition} Decision rule induced by belief intervals. Let \({m}^{ \oplus  }\left( {S \mid  x}\right)\) denote the final fused mass function obtained by the proposed hybrid combination rule, and let \({Bel}\left( {\cdot  \mid  x}\right)\) and \({Pl}\left( {\cdot  \mid  x}\right)\) be the standard belief and plausibility functions induced by \({m}^{ \oplus  }\left( {S \mid  x}\right)\) . The decision score for each singleton hypothesis \({S}_{i}\) is defined as \label{def19} \end{definition}

\begin{equation}\tag{27} \label{eq:27}\begin{split}
\operatorname{Pti}\left( {{S}_{i} \mid  x}\right) &= \left( {1 - \frac{\operatorname{Pl}\left( {{S}_{i} \mid  x}\right)  - \operatorname{Bel}\left( {{S}_{i} \mid  x}\right) }{Pl_{\max} - Bel_{\min}}}\right) \\
&\quad \times \left( {\operatorname{Pl}\left( {{S}_{i} \mid  x}\right)  - \operatorname{Bel}\left( {{S}_{i} \mid  x}\right) }\right) + \left( {\operatorname{Bel}\left( {{S}_{i} \mid  x}\right)  - Bel_{\min}}\right)
\end{split}\end{equation}

and the final decision is made by

\begin{equation}\tag{28} \label{eq:28} \widehat{y}\left( x\right)  = \arg \mathop{\max }\limits_{{{S}_{i} = \left\{  {\theta }_{i}\right\}  }}\operatorname{Pti}\left( {{S}_{i} \mid  x}\right) . \end{equation}

This decision rule provides an interval-aware transformation. Specifically, \({Be}{l}_{\min }\) and \(P{l}_{\max }\) denote the minimum credibility and the maximum plausibility among all singleton hypotheses respectively. Global spread \(P{l}_{\max } - {Be}{l}_{\min }\) serve as a common reference scale for the current fused evidence. The coefficient \(1 - \left( {{Pl}\left( {{S}_{i} \mid  x}\right)  - {Bel}\left( {{S}_{i} \mid  x}\right) }\right) /\left( {P{l}_{\max } - {Be}{l}_{\min }}\right)\) is then introduced as a stability indicator of the credibility interval: it decreases monotonically with the interval width \({Pl}\left( {{S}_{i} \mid  x}\right)  - {Bel}\left( {{S}_{i} \mid  x}\right)\) , thereby penalizing hypotheses whose support is highly ambiguous or weakly concentrated.

Accordingly, \(\operatorname{Pti}\left( {S}_{i}\right)\) combines two complementary contributions. The term \({Bel}\left( {{S}_{i} \mid  x}\right)  - {Be}{l}_{\min }\) quantifies the relative advantage of the lower bound, reflecting the amount of commitment that is guaranteed for \({S}_{i}\) beyond the weakest supported hypothesis. The second term, $\left(1 - \frac{{Pl}({S}_{i} \mid x) - {Bel}({S}_{i} \mid x)}{Pl_{\max} - Bel_{\min}}\right) \cdot ({Pl}({S}_{i} \mid x) - {Bel}({S}_{i} \mid x))$, incorporates the potential support upper bound while adaptively discounting it by the interval stability. This design prevents overly optimistic decisions driven by large plausibility values that are accompanied by wide uncertainty intervals, and at the same time preserves the informative role of plausibility when the evidence is consistent and well-concentrated. As a result, the proposed score yields a robust Bayesian-style decision criterion that remains well-defined in the presence of non-specific focal elements, without forcing an artificial redistribution of NFE mass onto singleton hypotheses.

\section{Algorithmic design}\label{sec:4_algorithmic_design}

This section presents the complete algorithmic workflow of the proposed framework. The procedure consists of two distinct phases, namely offline training on historical experience (Algorithm~\ref{alg:offline}) and online inference (Algorithm~\ref{alg:online}), which correspond to the learning and utilization of historical experience, respectively.

\begin{algorithm}
\label{alg:offline}
\caption{Offline Training: Historical Experience Modeling}
\KwIn{Historical dataset $\mathcal{H} = \{(x_t, y_t)\}_{t=1}^{N}$; evidence sources $M = \{m_1, \ldots, m_h\}$ defined on FoD $\Theta$; sensitivity parameters $\eta, \gamma$}
\KwOut{Context-specific historical experience weights $\{\widetilde{JRW}_i^{c}\}_{i=1}^{h, \, c \in \mathcal{C}}$}
\tcc{Phase 1: Context partition via spectral clustering}
Construct BPA feature matrix $\mathbf{B} \in \mathbb{R}^{N \times (h \cdot |2^{\Theta}|)}$ from all evidence sources\;
Perform spectral clustering on $\mathbf{B}$ to obtain context partition $\mathcal{C} = \{c_1, \ldots, c_L\}$\;
\tcc{Phase 2: Regret--rejoice computation on historical samples}
\ForEach{sample $(x_t, y_t) \in \mathcal{D}$}{
    Fuse all evidence via Dempster's rule: $m^{\oplus}(\cdot \mid x_t) \leftarrow m_1 \oplus \cdots \oplus m_h$\;
    $\hat{y}_t \leftarrow \arg\max_{\theta_k \in \Theta} Bel_{m^{\oplus}}(\theta_k \mid x_t)$\;
    \lIf{$\hat{y}_t = y_t$}{$\delta_t \leftarrow 0$ \tcp*[f]{correct decision, skip}}
    \lElse{$\delta_t \leftarrow 1$}
    \If{$\delta_t = 1$}{
        Identify optimal evidence: $m_{\mathrm{best}} \leftarrow \arg\min_{m_i \in M} \widehat{K}(m_i, p_{\mathrm{target}})$ \tcp*{Eq.~\ref{eq:18}}
        Compute rejoice value: $j(m_{\mathrm{best}}) \leftarrow 1 - \exp\!\bigl(-\eta\, \widehat{K}(m_{\mathrm{best}}, m^{\oplus})\bigr)$ \tcp*{Eq.~\ref{eq:20}}
        \ForEach{candidate erroneous evidence $m_e \in M$}{
            Compute error-removal combination $m^{\oplus, -e}$ via Eq.~\ref{eq:19}\;
            Compute regret value: $r(m_e) \leftarrow 1 - \exp\!\bigl(-\gamma\, \widehat{K}(m^{\oplus, -e}, m^{\oplus})\bigr)$ \tcp*{Eq.~\ref{eq:21}}
        }
    }
}
\tcc{Phase 3: Aggregate and normalize historical weights}
\ForEach{context $c \in \mathcal{C}$}{
    \ForEach{evidence source $m_i \in M$}{
        $JRW_i^{c} \leftarrow \displaystyle\sum_{\substack{t=1 \\ \delta_t=1,\, c(x_t)=c}}^{N} j_t(m_i) - \sum_{\substack{t=1 \\ \delta_t=1,\, c(x_t)=c}}^{N} r_t(m_i)$ \tcp*{Eq.~\ref{eq:23}}
    }
    Normalize: $\widetilde{JRW}_i^{c} \leftarrow \dfrac{\exp(JRW_i^{c})}{\sum_{j=1}^{h} \exp(JRW_j^{c})}$ for all $i$ \tcp*{Eq.~\ref{eq:24}}
}
\Return{$\{\widetilde{JRW}{}_i^{c}\}_{i=1}^{h, \, c \in \mathcal{C}}$}
\end{algorithm}

In the training phase (historical experience modeling): given a historical dataset \(\mathcal{H} = {\left\{  \left( {x}_{t},{y}_{t}\right) \right\}  }_{t = 1}^{N}\) , SC is first performed on the BPA feature representations of individual samples to partition the heterogeneous decision space into \(L\) context scenarios \(C = \left\{  {{c}_{1},\ldots ,{c}_{L}}\right\}\) . For each historical sample \({x}_{t}\) , the BPA results provided by multiple evidence sources \(M = \left\{  {{m}_{1},\ldots ,{m}_{h}}\right\}\) are fused via Dempster's combination to obtain the aggregated result \({m}^{ \oplus  }\left( {\cdot  \mid  {x}_{t}}\right)\) . When the fused decision conflicts with the ground truth label \({y}_{t}\) , the regret-rejoice computation is triggered: the optimal evidence \({m}_{\text{ best }}\) is identified according to Eq.~\ref{eq:18}, and its rejoice value \(j\left( {m}_{\text{ best }}\right)\) is calculated using Eq.~\ref{eq:20}; subsequently, for each candidate erroneous evidence \({m}_{e} \in  M\) , the regret value \(r\left( {m}_{e}\right)\) is computed as per Eq.~\ref{eq:21}. The cumulative historical score \({JR}{W}_{i}^{c}\) of each evidence within scenario \(c\) is defined as the algebraic sum of rejoice and regret, which is then normalized via softmax to yield the scenario-specific historical experience weight.

In the inference phase (evidence fusion and decision-making): for a test sample \(\mathrm{x}\) , its belonging cluster \(c\left( x\right)\) is first determined, and the corresponding historical experience weights \(\widetilde{{JR}{W}_i^{c(x)}}\) are retrieved. The CCM matrix of the current evidence set is computed to obtain the GCCD \(\widehat{K}\) . The historically weighted evidence \({m}_{w}\) is constructed according to Eq.~\ref{eq:25}, followed by the application of the hybrid combination rule given in Eq.~\ref{eq:26}, which adaptively adjusts the weight allocation between the conflict and consensus terms via \({e}^{-\widehat{K}}\) . The \({m}_{w}\) is then processed through the belief intervals interval decision rule defined in Eqs.~\ref{eq:27}-\ref{eq:28} to output the final class label.

\begin{algorithm}
\label{alg:online}
\caption{Online Inference: Robust Evidence Fusion and Decision}
\KwIn{Test sample $x$; evidence sources $M = \{m_1, \ldots, m_h\}$ on FoD $\Theta$; historical weights $\{\widetilde{JRW}_i^{c}\}$; sensitivity parameters $\eta, \gamma$}
\KwOut{Predicted class label $\hat{y}$}
Determine cluster membership: $c(x) \leftarrow$ assign $x$ to nearest context in $\mathcal{C}$\;
Retrieve historical weights $\{\widetilde{JRW}_i^{c(x)}\}_{i=1}^{h}$\;
\tcc{Step 1: Compute CCM and GCCD}
\ForEach{pair $(m_i, m_j)$, $1 \leq i < j \leq h$}{
    Compute association $k(m_i, m_j)$ via Eq.~\ref{eq:12}\;
    Compute similarity $S(m_i, m_j)$ via Eq.~\ref{eq:14}\;
    Compute chaos-conflict $\widehat{K}(m_i, m_j) \leftarrow 1 - S(m_i, m_j)$ \tcp*{Eq.~\ref{eq:ccm}}
}
$\widehat{K} \leftarrow \dfrac{2}{h(h-1)} \displaystyle\sum_{i < j} \widehat{K}(m_i, m_j)$ \tcp*{GCCD, Eq.~\ref{eq:16}}
\tcc{Step 2: Construct historical-experience weighted evidence}
$m_w(S \mid x) \leftarrow \displaystyle\sum_{i=1}^{h} \widetilde{JRW}_i^{c(x)} \cdot m_i(S \mid x), \quad \forall S \subseteq 2^{\Theta}$ \tcp*{Eq.~\ref{eq:25}}
\tcc{Step 3: Hybrid combination rule}
$\alpha \leftarrow 1 - e^{-\widehat{K}}$ \tcp*{conflict weight}
$\beta \leftarrow e^{-\widehat{K}}$ \tcp*{consensus weight}
$m_{\text{conj}}(S) \leftarrow \displaystyle\sum_{A_1 \cap \cdots \cap A_h = S \neq \emptyset} \prod_{i=1}^{h} m_i(A_i)$ \tcp*{conjunctive term}
$m_{\text{disj}}(S) \leftarrow \displaystyle\sum_{\substack{A_1 \cap \cdots \cap A_h = \emptyset \\ A_1 \cup \cdots \cup A_h = S}} \prod_{i=1}^{h} m_i(A_i)$ \tcp*{disjunctive term}
$m^{\oplus}(S) \leftarrow \alpha \cdot (m_{\text{conj}}(S) + m_{\text{disj}}(S)) + \beta \cdot m_w(S)$ \tcp*{Eq.~\ref{eq:26}}
\tcc{Step 4: Belief-interval decision}
\ForEach{singleton hypothesis $\theta_k \in \Theta$}{
    $Bel(\theta_k \mid x) \leftarrow \displaystyle\sum_{A \subseteq \{\theta_k\}} m^{\oplus}(A \mid x)$\;
    $Pl(\theta_k \mid x) \leftarrow \displaystyle\sum_{A \cap \{\theta_k\} \neq \emptyset} m^{\oplus}(A \mid x)$\;
}
$Bel_{\min} \leftarrow \min_{\theta_k} Bel(\theta_k \mid x)$;\quad $Pl_{\max} \leftarrow \max_{\theta_k} Pl(\theta_k \mid x)$\;
\ForEach{$\theta_k \in \Theta$}{
    $Pti(\theta_k \mid x) \leftarrow \left(1 - \dfrac{Pl(\theta_k \mid x) - Bel(\theta_k \mid x)}{Pl_{\max} - Bel_{\min}}\right)\!\left(Pl(\theta_k \mid x) - Bel(\theta_k \mid x)\right) + \left(Bel(\theta_k \mid x) - Bel_{\min}\right)$ \tcp*{Eq.~\ref{eq:27}}
}
$\hat{y} \leftarrow \arg\max_{\theta_k} Pti(\theta_k \mid x)$ \tcp*{Eq.~\ref{eq:28}}
\Return{$\hat{y}$}
\end{algorithm}

\section{Experiment}\label{sec:5_experiment}

\subsection{Experimental settings}

\subsubsection{Datasets}

Table~\ref{tab:experimental_datasets} summarises the 16 real world datasets used in the experiments. All were obtained from the UCI Machine Learning Repository and the NIH-CIP Repository, which collectively cover a broad spectrum of characteristics: high or low dimensional feature spaces, binary or multi class targets, and balanced or imbalanced distributions. The data covers fields such as biology, medicine, geography, and demography analysis. To prevent scale-dependent bias in the classifiers, every dataset was standardized before use.

\begin{table*}[t]
\centering
\caption{Experimental datasets}
\label{tab:experimental_datasets}
\adjustbox{max width=\textwidth}{
\begin{tabular}{cccc}
\toprule
Datasets & Sample & Class (Ratio) & Feature (Continuous/Discrete) \\
\cmidrule{1-4}
Accent & 329 & 6(0.5:0.14:0.09:0.09:0.09:0.09) & 12(12/0) \\
BMNP & 63 & 3(0.62:0.25:0.13) & 36(16/20) \\
Diabetes & 520 & 2(0.62:0.38) & 16(1/15) \\
Forest & 523 & 4(0.37:0.3:0.16:0.16) & 27(27/0) \\
German & 1000 & 2(0.7:0.3) & 24(3/21) \\
Hfcr & 299 & 2(0.68:0.32) & 12(8/4) \\
Ionosphere & 351 & 2(0.64:0.36) & 34(32/2) \\
ISPY & 215 & 2(0.73:0.27) & 18(4/14) \\
Sonar & 208 & 2(0.53:0.47) & 60(60/0) \\
Sports & 1000 & 2(0.64:0.36) & 59(49/10) \\
Turkish & 400 & 4(0.25:0.25:0.25:0.25) & 50(50/0) \\
Vehicle & 846 & 4(0.26:0.26:0.25:0.24) & 18(18/0) \\
Wine & 178 & 3(0.33:0.4:0.27) & 13(13/0) \\
WDBC & 569 & 2(0.63:0.37) & 30(30/0) \\
WPBC & 198 & 2(0.76:0.24) & 33(31/2) \\
Zoo & 101 & 7(0.41:0.2:0.13:0.1:0.08:0.05:0.04) & 16(0/16) \\
\bottomrule
\end{tabular}
}
\end{table*}

\subsubsection{Baseline and hyperparameter settings}

To demonstrate the superior inference capability of the proposed framework, we benchmark it against two representative classes of methods: evidential reasoning and ensemble learning. Within the evidential reasoning category, the comparative methods include Dempster's rule \citep{dempster_upper_1967}, Yager's rule \citep{yager_dempster-shafer_1987}, Dubois' rule \citep{dubois_representation_1988}, Murphy's averaging rule \citep{murphy_combining_2000}, Deng's weighted rule \citep{yong_combining_2004}, Belief Hellinger Fusion method (BHF) \citep{zhu_belief_2021}, EC-FMDS \citep{chamlal_ensemble_2025}, and the Hierarchical Evidence Fusion method (HEF) \citep{Zhang_2025}. For ensemble learning, comprehensive comparative experiments are conducted against three gradient boosting methods, namely eXtreme Gradient Boosting (XGBoost) \citep{chen_xgboost_2016}, Light Gradient Boosting Machine (LightGBM) \citep{ke_lightgbm_2017}, and Categorical Boosting (CatBoost) \citep{prokhorenkova_catboost_2018}.

To ensure a fair comparison, given that the ensemble learning methods utilize decision trees (DTs) as base learners, we similarly employ DTs as evidence bodies within the DST framework. Random split selection is adopted to introduce necessary stochasticity into each evidence body (i.e., each individual DT), while hyperparameters such as tree depth are configured according to recommendations to promote generalization \citep{dhanka_advancements_2026}. During the comparative experiments, the number of base DTs in the ensemble methods is set equal to the number of evidence bodies in the DST methods. Detailed parameter settings are provided in Table~\ref{tab:hyperparameter_settings}; parameters not listed in the table follow the default values of the official Python libraries.

\begin{table*}[t]
\centering
\caption{Hyperparameter settings}
\label{tab:hyperparameter_settings}
\adjustbox{max width=\textwidth}{
\begin{tabular}{ccc}
\toprule
Type & Methods & Hyperparameter settings \\
\cmidrule{1-3}
Evidence & DT & splitter: random, max\_depth: 8, min\_samples\_leaf: 2, min\_samples\_split: 4, max\_features: "sqrt", class\_weight: "balanced" \\
\cmidrule{2-3}
Cluster & SC & n\_clusters: number of categories, affinity: rbf, gamma: 1, assign\_labels: kmeans, eigen\_solver:arpack, n\_init: 20 \\
\cmidrule{2-3}
\multirow{3}{*}{Ensemble learning} & CatBoost & learning\_rate: 0.1, depth: 8, l2\_leaf\_reg: 1 \\
 & LightGBM & learning\_rate: 0.1, max\_depth: 8, subsample: 1.0, colsample\_bytree: 1.0 \\
 & XGBoost & learning\_rate: 0.1, max\_depth: 8, subsample: 1.0, colsample\_bytree: 1.0, reg\_lambda:1e-5 \\
\cmidrule{2-3}
\multirow{10}{*}{DST} & Dempster & - \\
 & Deng & Distance: Jousselme \\
 & Dubois & - \\
 & EC-FMDS & q=2 \\
 & HEF & - \\
 & Murphy & - \\
 & Yager & - \\
 & BHF & - \\
 & Our & $\eta$: 0.5, $\gamma$: 0.5, $|C|$: number of class \\
\bottomrule
\end{tabular}
}
\end{table*}

\subsubsection{Procedure and metrics}

All experiments were conducted on a system with an Intel Core Ultra 5 225H CPU, 32 GB RAM, and Windows 11 25H2. The code was implemented in Python 3.9.25, and all stochastic procedures were fixed with a random seed of 42 to ensure reproducibility.

To ensure fair and comprehensive evaluation, all methods were assessed under a unified experimental protocol across the 16 datasets listed in Table~\ref{tab:experimental_datasets}. Each dataset was randomly partitioned using 5-fold cross-validation. Reported results are presented as mean values, standard deviations, and rankings (where applicable) across all folds to ensure statistical reliability and robustness.

To further validate the proposed method, we conducted additional analyses encompassing robustness evaluation under noisy conditions, adjustment of the number of evidence bodies, and substitution of the evidence body base model, together with sensitivity analysis and ablation studies of hyperparameters. These complementary experiments provide deeper insights into the stability, generalization capability, and component contributions of the proposed framework. Detailed experimental procedures are presented in the corresponding sections below.

For comprehensive assessment of classification performance, we employed four widely adopted evaluation metrics: accuracy (ACC), precision (PRE), recall (REC), and F1-score (F1). ACC measures the overall proportion of correctly classified samples, whereas PRE and REC evaluate the correctness and completeness of positive predictions, respectively. The F1, defined as the harmonic mean of precision and recall, offers a balanced assessment particularly suited to imbalanced datasets. We adopted macro averaging for all metrics. Additionally, receiver operating characteristic (ROC) curves and the area under the curve (AUC) were employed to evaluate the discriminative capability of different methods. The predictions from all folds were pooled to generate a single ROC curve and calculate the AUC.

\subsection{Results}\label{sec:52}

\subsubsection{Comparative analysis}\label{sec:521}

This section uniformly employed three decision trees as evidence sources to construct the DST framework, with the number of base learners matching that of the compared ensemble methods. Table~\ref{tab:performance_comparison_mean_pm_std_of_different_methods_with_average_rank} summarizes the mean performance, standard deviation, and average rank of each method in terms of ACC, PRE, REC, and F1. Our method achieved optimal mean values across all evaluation metrics (ACC: 87.01, PRE: 85.44, REC: 87.51, F1: 85.78), with average ranks of 3.31, 3.52, 2.89, and 3.22, respectively. Detailed performance results for each dataset are provided in Appendix A.

Compared with the DST model, the proposed framework demonstrates superior capability in quantifying and resolving high-conflict evidence. The performance degradation of Dempster's rule (ACC: 80.30, F1: 76.03) stems from its rigid normalization mechanism, which tends to produce counterintuitive fusion results. Yager's approach reallocates conflict mass to the frame of discernment; however, this conservative strategy indiscriminately discards valuable informative discrimination signals, manifesting as the second-lowest F1 score among all methods (74.87). Murphy's average rule enhances stability through uniform evidence weighting, yet its assumption of source consistency fails to account for context-dependent reliability variations; despite its smoothing effect, it shows no substantial advantage over Dempster's rule (F1: 77.74 vs. 76.03), indicating inadequate handling of such uncertainty. Deng's method introduces source differentiation through Jousselme distance-based weighting, but measures only static BPA discrepancies without modeling the internal uncertainty structure of FEs, limiting its adaptability under epistemic uncertainty. The BHF method employs belief Hellinger distance to quantify evidence dissimilarity and incorporates entropy to characterize ambiguity, thereby establishing a weighted average fusion strategy. However, its weight generation relies exclusively on inter-evidence distance metrics and intrinsic uncertainty, without leveraging historical decision feedback; consequently, it fails to capture reliability variations of evidence sources across diverse decision contexts. EC-FMDS attempts to integrate feature selection preprocessing with fusion mechanisms, yet their independent operational modes cannot resolve conflicts arising from evidence-level inconsistency, as corroborated by its substantial performance ranking variance. HEF's hierarchical structure mitigates conflict and reduces computational complexity through averaged evidence sources; however, its strict fusion sequence and evidence averaging mechanism may lead to premature determination of unreliable averaged evidence system combinations when evidence bodies are scarce and homogeneous (e.g., three decision trees in this experiment), evidenced by its lowest F1 (75.36).

Compared with gradient boosting frameworks, the proposed method exhibits marked superiority in F1 score (85.78 versus 75.05, 77.87, and 78.35). Although all three ensemble methods construct additive models through gradient optimization to improve fitting accuracy by minimizing empirical risk, they lack explicit modeling of inter-evidence uncertainty. In contrast, the proposed framework characterizes decision boundary uncertainty through belief and plausibility functions, thereby preserving more discriminative information in regions with imbalanced class distributions or overlapping feature spaces and achieving superior predictive performance.

\begin{figure*}[!t]
\centering
\includegraphics[max width=0.9\textwidth]{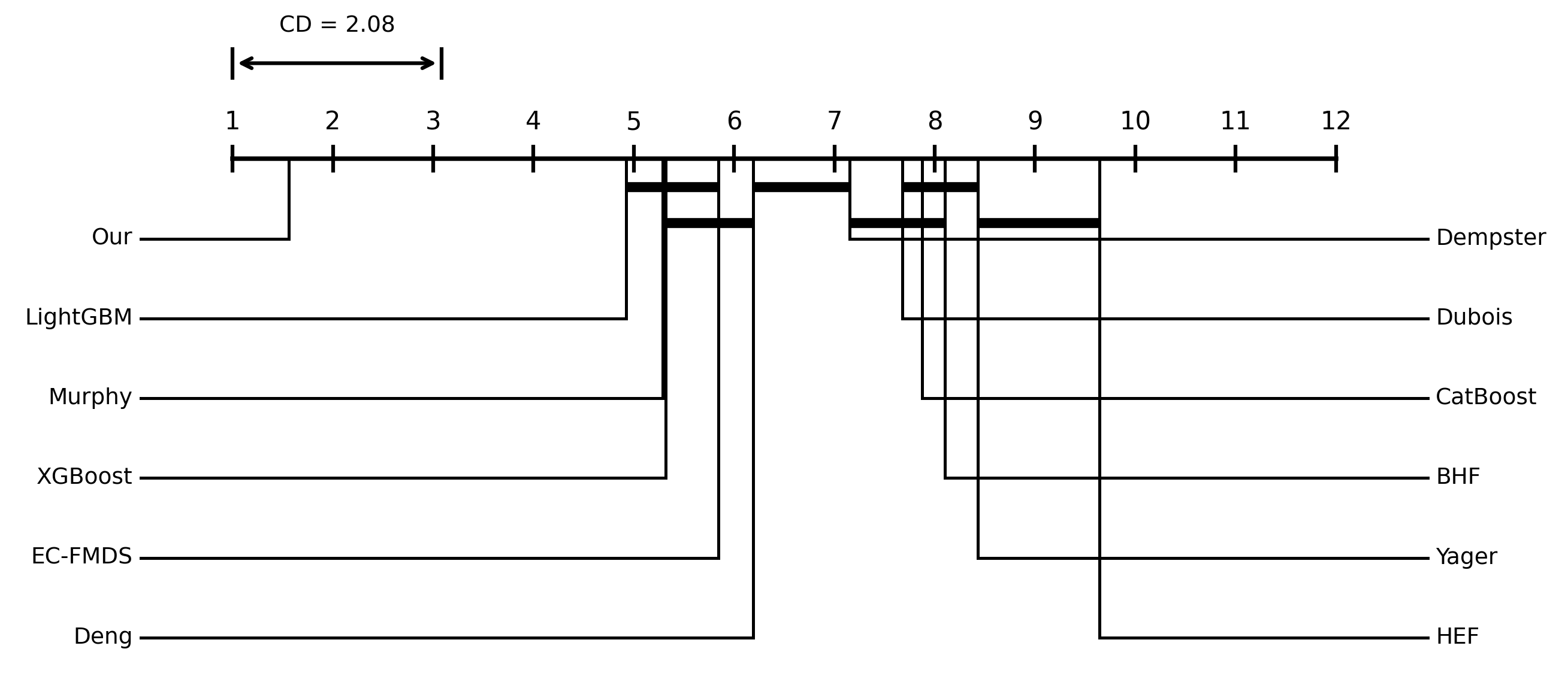}
\caption{CD diagram of all models.}
\label{fig:31_267_644_1268_541_0}
\end{figure*}

The proposed framework overcomes these limitations through a unified CCM mechanism. CCM simultaneously quantifies cross-evidence conflict and intra-evidence non-specificity without presupposing fusion order or structural assumptions; furthermore, it introduces historical experience-driven weighting based on regret theory, enabling evidence reliability assessment to adapt dynamically to current sample conditions and maintaining fine-grained reliability discrimination even under evidence-scarce scenarios. This end-to-end design avoids the decoupling of preprocessing and fusion, achieving collaborative optimization of evidence quality assessment and fusion decision. According to the Friedman test results and Nemenyi post-hoc test (CD = 2.08) presented in Figure~\ref{fig:31_267_644_1268_541_0}, the proposed framework achieved the highest average rank among all compared methods, demonstrating statistically significant performance advantages over every alternative approach. Among the competing methods, LightGBM, Murphy, XGBoost, EC-FMDS, and Deng formed a relatively competitive cluster with no statistically significant differences detected within this group, whereas Dempster, Dubois, CatBoost, BHF, Yager, and HEF occupied lower overall rankings.

\begin{table*}[t]
\centering
\caption{Performance comparison (mean \(\pm\) std) of different methods with average rank}
\label{tab:performance_comparison_mean_pm_std_of_different_methods_with_average_rank}
\adjustbox{max width=\textwidth}{
\begin{tabular}{cccccccccc}
\toprule
Type & Model & ACC & ACC\_rank & PRE & PRE\_rank & REC & REC\_rank & F1 & F1\_rank\\\cmidrule{1-10}
\multirow{3}{*}{Ensemble learning} & CatBoost & 80.21±13.30 & 7.69±4.78 & 77.36±17.02 & 8.44±4.76 & 74.93±18.11 & 10.52±4.58 & 75.05±18.25 & 9.94±4.68\\\multirow{2}{*}{\phantom{X}}  &  LightGBM  & 83.10±9.90 &  4.39±3.32  &  79.08±15.41  &  4.95±3.53  &  78.00±15.00  &  5.53±3.60  &  77.87±15.54  &  5.28±3.58\\ &  XGBoost  & 82.26±11.20 &  4.97±2.98  &  79.15±14.50  &  5.48±3.03  &  78.48±14.74  &  5.50±3.20  &  78.35±14.83  &  5.36±3.15\\\cmidrule{2-10}
\multirow{9}{*}{DST} & Dempster & 80.30±12.90 & 8.64±4.16 & 77.44±17.30 & 9.72±4.45 & 76.61±16.80 & 9.84±4.15 & 76.03±17.56 & 10.09±4.13\\ &  Deng  & 80.44±13.55 &  8.11±4.25  &  77.50±16.02  &  9.31±4.65  &  77.66±16.06  &  8.70±4.36  &  77.22±16.19  &  8.92±4.38\\ &  Dubois  & 80.18±12.57 &  8.52±4.07  &  76.60±15.91  &  10.11±4.05  &  76.85±15.76  &  9.44±4.07  &  76.20±16.06  &  9.58±3.99\\ &  EC-FMDS  & 80.31±14.06 &  5.48±3.01  &  77.55±16.13  &  6.08±3.24  &  77.79±16.09  &  5.41±3.19  &  77.14±16.36  &  5.62±3.25\\ &  HEF  & 78.51±13.33 &  9.98±4.35  &  75.36±16.36  &  10.64±4.10  &  74.36±16.54  &  11.33±3.83  &  73.63±17.15  &  11.53±3.77\\ &  Murphy  & 80.89±13.97 &  5.48±3.26  &  78.35±16.06  &  5.89±3.26  &  78.11±16.56  &  5.47±2.97  &  77.74±16.61  &  5.55±3.01\\ &  Yager  & 78.80±13.54 &  7.56±3.56  &  75.65±16.59  &  7.70±3.67  &  75.45±16.31  &  7.58±3.58  &  74.87±16.86  &  7.83±3.51\\ &  BHF  & 79.34±13.84 &  9.08±4.04  &  75.89±17.12  &  10.34±3.88  &  76.80±16.55  &  9.38±3.83  &  75.87±16.95  &  9.88±3.92\\ &  Our  &  \textbf{87.01±13.93}  &  \textbf{3.31±3.78}  &  \textbf{85.44±15.29}  &  \textbf{3.52±3.90}  & \textbf{87.51±14.75} &  \textbf{2.89±3.51}  &  \textbf{85.78±15.42} & \textbf{3.22±3.77}\\
\bottomrule
\end{tabular}
}
\end{table*}

\begin{figure*}[!t]
\centering
\includegraphics[max width=1.0\textwidth]{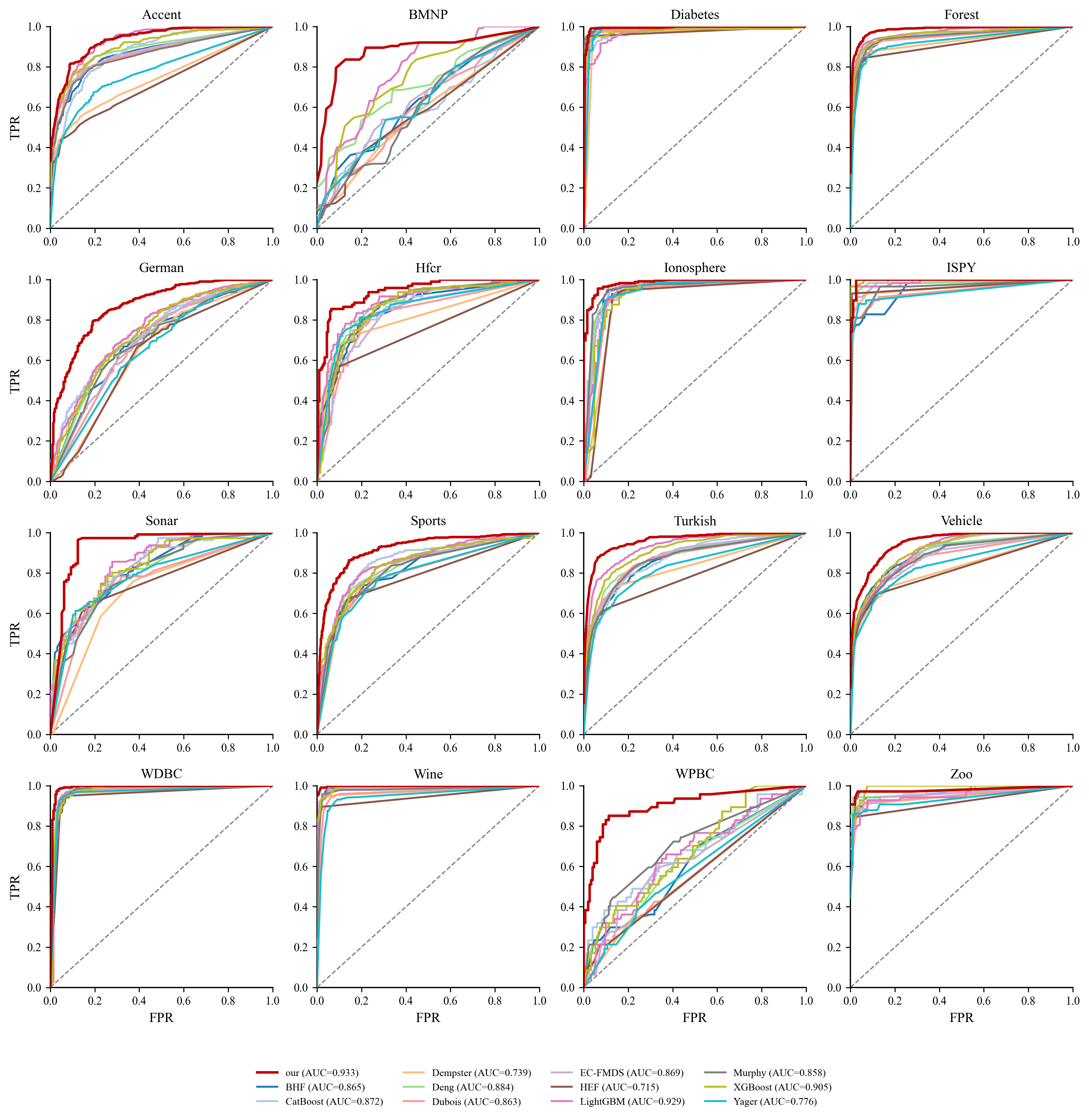}
\caption{Comparison of ROC curves for different models on 16 datasets}
\label{fig:32_262_328_1275_1300_0}
\end{figure*}

Figure~\ref{fig:32_262_328_1275_1300_0} presents the ROC curves for all methods across the 16 datasets. The proposed method (red curve) consistently occupies the uppermost position in the majority of cases, exhibiting substantially higher true positive rates at low false positive rates. This indicates superior discriminative capability and well-calibrated uncertainty estimation, particularly in scenarios characterized by ambiguous decision boundaries or class imbalance.

Closer examination reveals two distinct patterns. On datasets with inherently high separability, such as Diabetes, ISPY, and Wine, performance differences among methods diminish considerably, suggesting that simple evidence aggregation strategies suffice when classification tasks are relatively straightforward. Conversely, on challenging datasets with substantial class overlap, including BMNP, German, and WPBC, the proposed framework demonstrates marked performance advantages over competing approaches. This pronounced separation in difficult scenarios underscores the efficacy of our uncertainty-aware evidence fusion mechanism. Consequently, the proposed framework achieves the highest overall mean AUC of 93.3, with its robust generalization across diverse datasets attributable primarily to exceptional performance in discriminatively demanding conditions rather than incremental gains on already well-resolved tasks.

\subsubsection{Robustness analysis}

To verify the robustness of the proposed framework under diverse conditions, we conducted a series of experiments, which included introducing data noise and varying both the types and the number of evidence sources.

\fourlevelsection{Noise}

To comprehensively evaluate the robustness of the proposed framework under noisy conditions, we introduced multiple types of data noise during the training phase to emulate interference scenarios commonly encountered in real-world applications. Specifically, feature-level noise, label-level noise, and their combinations as hybrid noise were considered to assess the stability and generalization capability of the framework across varying noise intensities.

For feature-level perturbations, two noise injection strategies were adopted. The first randomly replaced a proportion of feature values with the corresponding feature means, thereby modeling data missingness during the acquisition process. The second randomly added noise sampled from a zero-mean Gaussian distribution to selected features, simulating stochastic interference with a normal distribution. For label-level noise, a specified ratio of sample labels was randomly reassigned to alternative classes, reflecting typical annotation errors in practical datasets.

To further examine the compound effects of concurrent feature and label corruption, two hybrid noise configurations were designed: uniform feature noise combined with random label noise, and Gaussian-distributed feature noise combined with random label noise. Noise intensities were set to 5\%, 10\%, 15\%, and 20\%, enabling a systematic analysis of performance degradation as noise levels increased.

\begin{figure*}[!t]
\centering
\includegraphics[max width=0.9\textwidth]{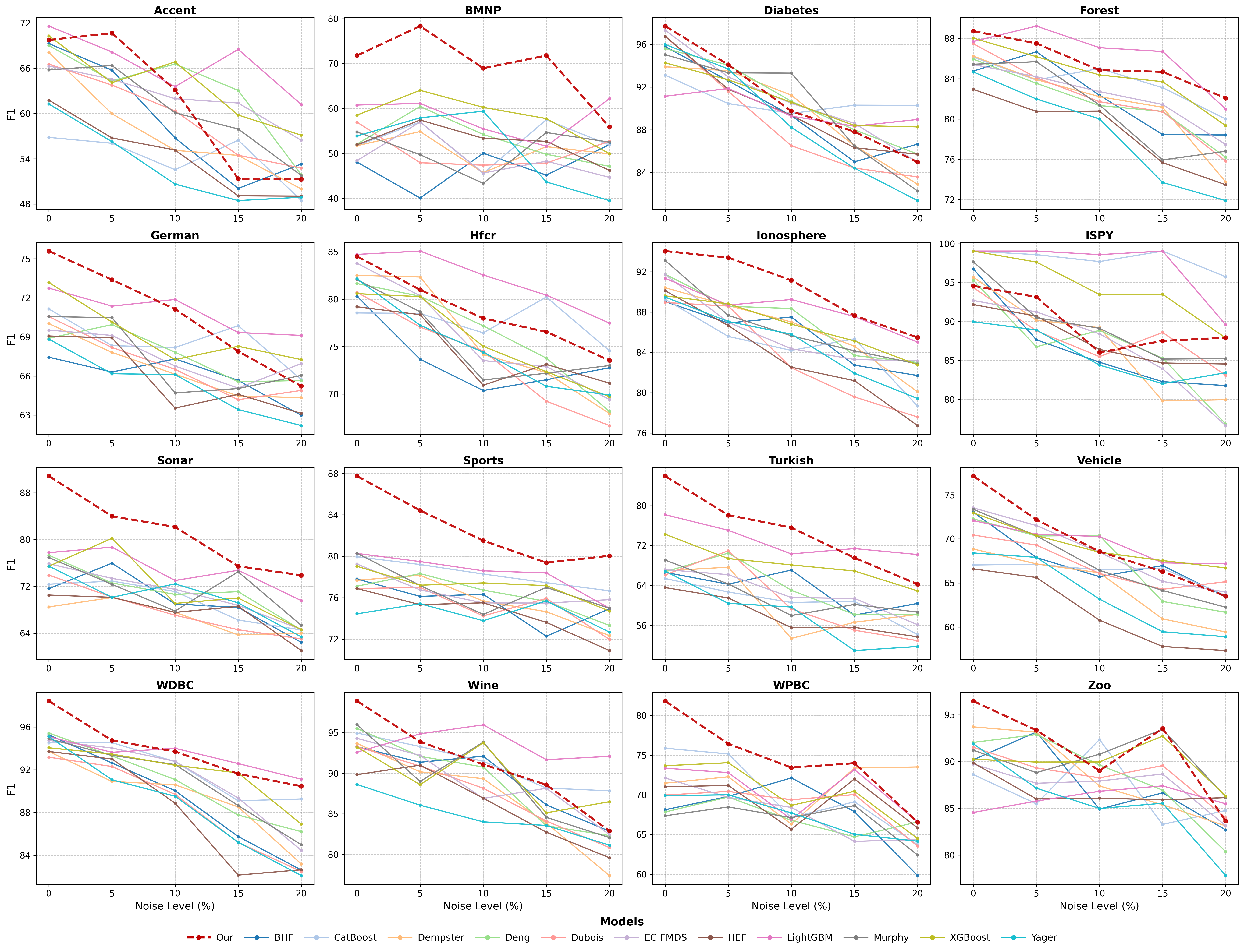}
\caption{F1-score under hybrid noise (gaussian feature noise and random label)}
\label{fig:34_315_196_1169_892_0}
\end{figure*}

\begin{figure*}[!t]
\centering
\includegraphics[max width=0.9\textwidth]{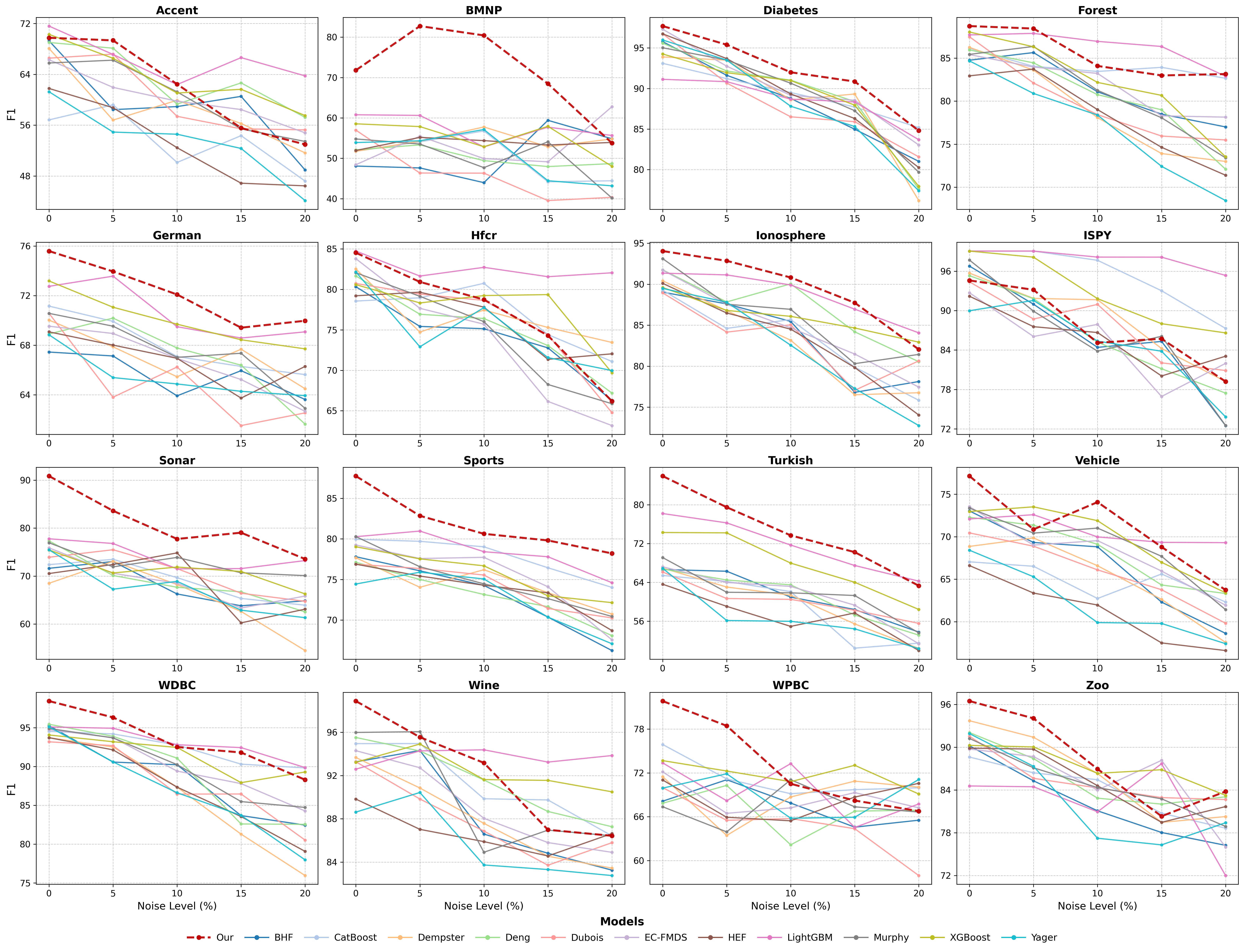}
\caption{F1-score under hybrid noise (mean feature noise and random label)}
\label{fig:34_319_1154_1162_880_0}
\end{figure*}

Figure~\ref{fig:34_315_196_1169_892_0} and Figure~\ref{fig:34_319_1154_1162_880_0} present the variations in F1 performance of the compared methods under two mixed-noise scenarios, while \hyperlink{app:B}{Appendix B} presents the corresponding F1 results under the three basic noise settings. Overall, as the noise level gradually increases from 5\% to 20\%, the performance of all methods declines to varying degrees. Nevertheless, the proposed method maintains a relatively high F1 score under most noise settings. This result indicates that the proposed approach is not effective only in relatively ideal data environments; rather, it can still sustain stable decision performance when the training samples are affected by feature contamination, label shifts, or their combined interference.

In contrast to most existing DST-based methods, which typically rely on static processing according to the differences among the current BPAs, they struggle to distinguish anomalous evidence caused by incidental noise from valid evidence with long-term stability. In specific contexts, the proposed method exhibits greater robustness under noise. On the one hand, it employs an adaptive dual-weighting mechanism based on historical experience and current conflict. On the other hand, its hybrid combination rule preserves uncertain mass instead of forcibly compressing it into the SFE, thereby reducing the dominance of anomalous evidence and the associated noise in the final fusion result while maintaining relatively stable performance.

It should nevertheless be acknowledged that, compared with ensemble tree-based methods, most DST-based approaches, including the proposed method in some cases, are generally less stable under noisy conditions. This difference is directly related to the distinct modeling priorities of the two methodological paradigms. Ensemble tree methods are fundamentally developed within a supervised learning framework, where classification errors are continuously fitted through iterative empirical risk optimization or the parallel integration of multiple heterogeneous tree structures, and their objective functions are therefore closely aligned with the final decision outcome \citep{van_Engelen_2019,Alzubaidi_2021}. Ensemble strategies and recursive splitting will mitigate the instability of individual trees and promote complementary effects across learners \citep{Hasan_2024}. By contrast, the main strength of DST methods lies in the explicit representation of uncertainty and the management of conflict, rather than in the direct minimization of classification error. The fusion process in evidence theory is typically centered on BPA construction and combination rule design, with the primary emphasis placed on integrating support relationships among multiple evidence sources. However, this process does not inherently guarantee sufficiently strong discriminative learning of class boundaries, and fusion rules mainly redistribute information under existing quality constraints rather than iteratively correcting prediction bias to improve generalization, as ensemble tree models do \citep{Qiao_2023b}. For this reason, the proposed method further incorporates historical experience modeling and conflict-adaptive fusion within the DST framework, thereby compensating for the limited discriminative learning capability of conventional evidence fusion methods and enabling it to narrow the gap with, or even outperform, ensemble models under noisy conditions.

Interestingly, under mixed-noise scenarios, the performance of most methods declines relatively gradually as noise intensity increases, whereas under single-noise conditions, performance exhibits more pronounced fluctuations and irregularity. This observation suggests that perturbations do not necessarily exacerbate model degradation in a purely monotonic manner; rather, they may encourage the formation of more robust decision boundaries, thereby improving adaptability to distribution shifts and indirectly enhancing generalization. This finding is consistent with the conclusions of \citet{liu_synergy_2023}, who reported that moderate noise exposure helps mitigate overfitting and improves model stability in complex and uncertain environments.

\begin{figure*}[!t]
\centering
\includegraphics[max width=1.0\textwidth]{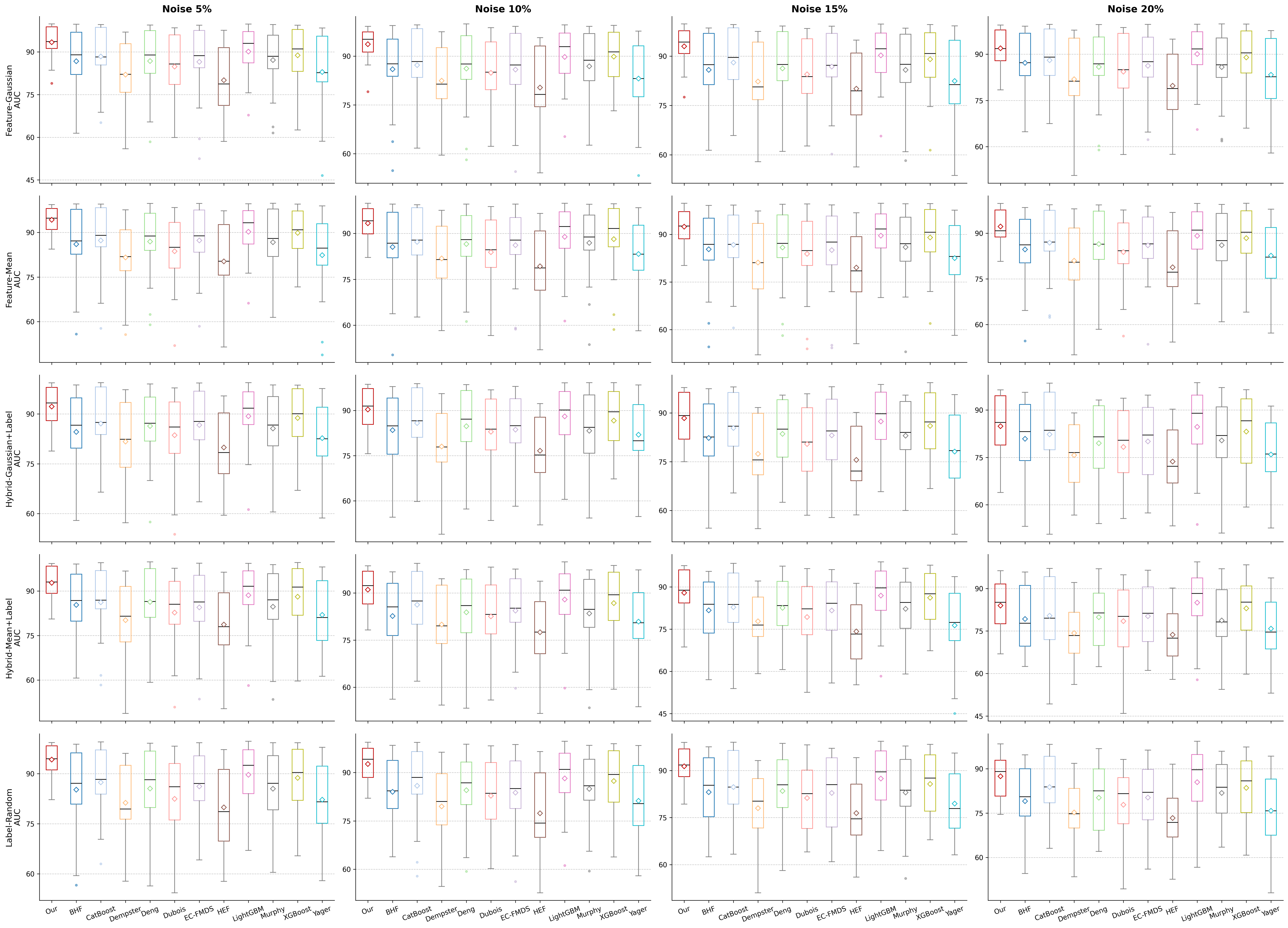}
\caption{Boxplots of average AUC for different models under various noise types and levels}
\label{fig:37_252_186_1292_946_0}
\end{figure*}

Figure~\ref{fig:37_252_186_1292_946_0} further illustrates the distribution of AUC values for different methods across varying noise types and intensities, providing a complementary perspective on robustness from the standpoint of ranking-based discriminative ability. Overall, as the noise level increases, the median AUC of most methods shifts downward and the interquartile range tends to widen, indicating that noise affects not only average discriminative performance but also cross-dataset stability. The proposed method maintains a relatively competitive AUC distribution under most noise settings, as evidenced by a higher median and mean, a narrower interquartile range, and fewer outliers (corresponding to markedly poor performance). This suggests that its advantages are not confined to threshold-dependent classification metrics, but also extend to sample-level discriminability. Figure~\ref{fig:37_252_186_1292_946_0} also shows that the boxplots of AUC obtained under mixed-noise conditions appear more scattered than those under single-noise conditions, which is broadly consistent with the pattern observed for F1.

\fourlevelsection{Different number of evidences}

\begin{figure*}[!t]
\centering
\includegraphics[max width=0.9\textwidth]{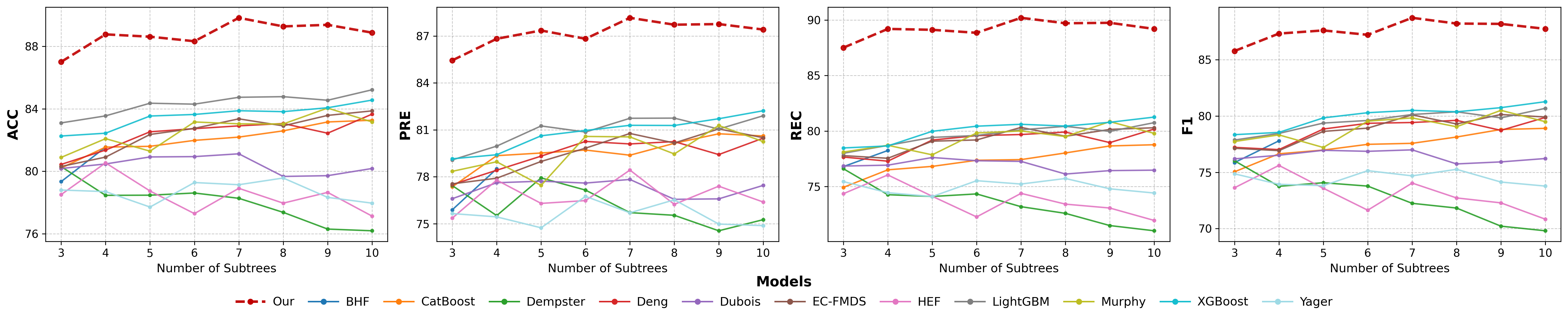}
\caption{Comparative performance of models across subtree sizes}
\label{fig:38_262_876_1273_262_0}
\end{figure*}

Figure~\ref{fig:38_262_876_1273_262_0} shows that as the number of basic evidences (trees) increases, the performance of most methods generally follows a pattern of improvement followed by stabilization. When the number of evidences is small, individual tree sources exhibit strong randomness, and the fusion result is more susceptible to incidental bias. As the number of evidences increases, both tree diversity and collective stability improve, leading to better classification performance. However, with further increases, the complementary information provided by additional trees gradually diminishes, and the performance gains tend to saturate. Certain approaches, such as Dempster and HEF, fail to handle the increased conflict effectively as the number of evidence sources grows, resulting in performance degradation. By contrast, the proposed method maintains consistently strong and more stable performance across different numbers of evidences, indicating its ability to balance evidence complementarity with conflict suppression effectively.

\fourlevelsection{Different base model}

To further examine the proposed framework's dependence on the evidence generators, this section keeps all other experimental settings unchanged and replaces the original combination of multiple random DTs used to construct the evidence bodies with a set of seven different base evidence generators: DT, Support Vector Classification (SVC), Logistic Regression (LR), Gaussian Process Classifier (GPC), K-Nearest Neighbors (KNN), Multilayer Perceptron (MLP), and Gaussian Naive Bayes (GNB). All models use the default parameters provided by scikit-learn, and the fusion results of different DST methods are compared accordingly.

Table~\ref{tab:performance_comparison_under_different_base_models_mean_pm_std_with_average_rank} presents the corresponding results. Compared with the relatively stable DT-based setting, replacing the base models introduces greater overall performance fluctuations, indicating that evidence quality, output diversity, and fusion compatibility all affect the final decision. Under this setting, the proposed framework achieves an ACC of 83.42, a PRE of 81.21, a REC of 78.34, and an F1 score of 78.43, with corresponding average ranks of 4.44, 4.91, 4.64, and 4.81. Overall, it remains competitive and outperforms most of the compared DST methods. Although the set of base evidence generators includes several relatively weak models, almost all DST methods achieve favorable performance after evidence fusion. Notably, the proposed framework maintains comparatively stable overall performance after the replacement of base models, as reflected by its lower standard deviations in both performance and ranking, demonstrating the framework's strong adaptive evidence fusion capability when confronted with heterogeneous base models of varying quality.

\begin{table*}[t]
\centering
\caption{Performance comparison under different base models (mean \(\pm\) std) with average rank}
\label{tab:performance_comparison_under_different_base_models_mean_pm_std_with_average_rank}
\adjustbox{max width=\textwidth}{
\begin{tabular}{cccccccccc}
\toprule
Type & Model & ACC & ACC\_rank & PRE & PRE\_rank & REC & REC\_rank & F1 & F1\_rank\\\cmidrule{1-10}
\multirow{7}{*}{Base evidence} & SVC & 76.81±17.45 & 9.25±5.71 & 76.48±16.14 & 8.98±5.09 & 75.21±16.85 & 8.00±5.43 & 74.34±17.69 & 8.22±5.46\\ &  LR  & 75.50±16.66 &  7.97±4.48  &  68.75±22.90  &  8.73±4.18  &  70.60±17.21  &  8.31±4.31  &  67.74±21.09  &  8.62±4.15\\ &  GPC  & 63.95±21.67 &  10.64±4.98  &  60.58±23.39  &  10.95±4.69  &  56.61±21.23  &  11.19±4.81  &  50.00±25.85  &  11.61±4.41\\ &  DT  & 78.20±13.42 &  9.34±5.55  &  74.27±16.59  &  9.75±5.39  &  74.39±15.96  &  9.06±5.45  &  73.93±16.41  &  8.95±5.39\\ &  KNN  & 75.00±14.10 &  9.67±4.16  &  64.22±22.77  &  10.69±4.58  &  66.10±18.17  &  10.80±4.35  &  63.36±20.76  &  10.97±4.26\\ &  MLP  & 63.95±21.67 &  10.64±4.98  &  60.58±23.39  &  10.95±4.69  &  56.61±21.23  &  11.19±4.81  &  50.00±25.85  &  11.61±4.41\\ &  GNB  & 80.32±13.44 &  7.58±5.02  &  78.30±15.16  &  7.69±4.71  &  75.88±16.08  &  7.69±4.98  &  76.20±15.97  &  7.55±4.87\\\cmidrule{2-10}
\multirow{9}{*}{DST} & Deng & 80.51±11.33 & 7.36±4.73 & 77.33±15.21 & 7.67±4.49 & 75.30±15.02 & 7.41±4.51 & 75.09±15.46 & 7.36±4.46\\ &  HEF  & 80.20±13.50 &  5.97±3.55  &  75.84±19.71  &  6.86±3.55  &  75.32±16.52  &  6.41±3.38  &  73.97±18.88  &  6.66±3.53\\ &  Dubois  & 78.22±15.74 &  5.94±3.77  &  72.91±22.17  &  6.95±3.50  &  73.25±16.85  &  6.52±3.39  &  71.23±20.31  &  6.88±3.51\\ &  Dempster  & 80.14±13.49 &  6.16±3.56  &  75.79±19.69  &  7.00±3.58  &  75.29±16.48  &  6.56±3.39  &  73.93±18.85  &  6.81±3.51\\ &  EC-FMDS  & 82.89±11.03 &  \textbf{4.33±2.52}  &  81.38±14.63  &  5.00±3.07  &  77.76±14.51  &  5.02±2.50  &  77.92±15.13  &  4.97±2.70\\ &  Murphy  &  78.08±15.29 &  6.45±4.43  &  74.02±20.00  &  6.05±3.86  &  72.59±17.23  &  6.88±4.14  &  71.29±19.61  &  6.59±4.24\\ &  BHF  &  81.72±11.10 &  5.52±3.02  &  \textbf{81.46±14.36} & 5.75±3.26  &  75.77±15.57  &  6.52±3.02  &  75.62±16.40  &  6.48±3.06\\ &  Yager  &  77.70±16.21 &  6.31±4.10  &  72.07±22.49  &  7.36±3.77  &  72.77±17.23  &  6.86±3.67  &  70.66±20.84  &  7.20±3.76\\ &  Our  & \textbf{83.42±10.13} &  4.44±3.06  &  81.21±13.71  &  \textbf{4.91±2.98}  &  \textbf{78.34±13.90} & \textbf{4.64±2.85}  &  \textbf{78.43±14.28} & \textbf{4.81±3.04}\\
\bottomrule
\end{tabular}
}
\end{table*}

\subsubsection{Parameter Sensitivity}

To evaluate the influence of key parameters on the performance of the proposed framework, this section conducts parameter sensitivity analysis from two perspectives: historical context partitioning and experience-weight generation. Specifically, all other hyperparameters are fixed while varying the number of clusters obtained by spectral clustering, and the resulting changes in ACC, PRE, REC, and F1 under different clustering granularities are examined to assess the sensitivity of historical experience modeling to the context partition scale. Subsequently, a joint grid search is performed over the two sensitivity coefficients, \(\eta\) and \(\gamma\) , in the regret-rejoice mechanism. By plotting F1 heatmaps and ACC three-dimensional surfaces across multiple datasets, we further investigate the stability of the experience-weight mechanism and the distribution of high-performing regions under different parameter combinations.

\begin{figure*}[!t]
\centering
\includegraphics[max width=1.0\textwidth]{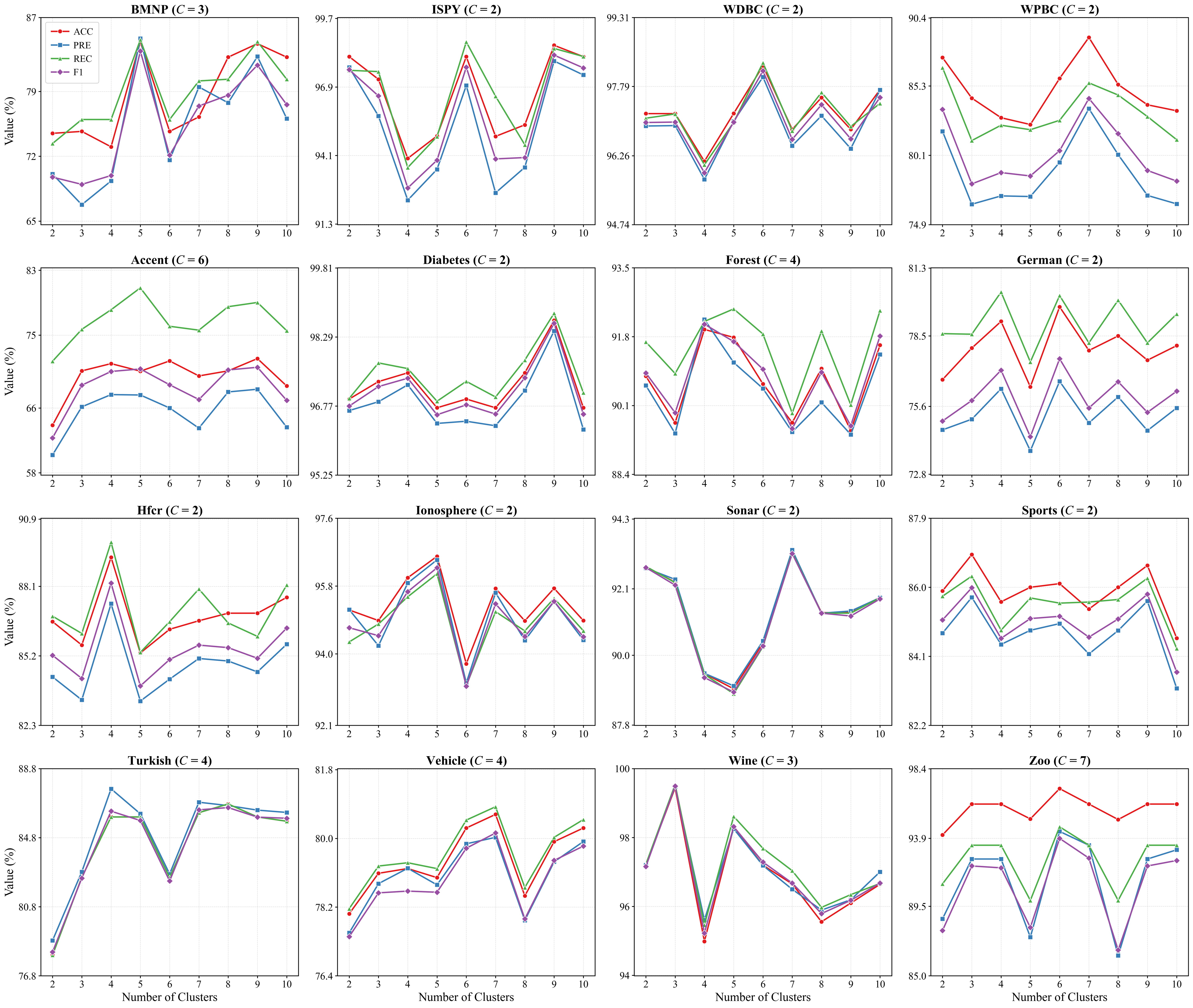}
\caption{Sensitivity analysis of clustering performance with respect to the number of clusters}
\label{fig:41_250_194_1298_1094_0}
\end{figure*}

Figure~\ref{fig:41_250_194_1298_1094_0} presents the performance variation of the proposed model under different numbers of clusters. Overall, model performance does not increase monotonically with the number of clusters, but instead exhibits dataset-dependent local optima associated with sample size, class structure, and feature dimensionality. For small-sample datasets such as BMNP, Wine, and Zoo, the performance curves fluctuate more noticeably, indicating that when limited samples are available for historical experience statistics, changes in the number of clusters directly affect the stability of the regret-rejoice values within each context. An overly fine-grained partition may lead to insufficient samples within individual clusters, making the estimation of experience weights more susceptible to local sample distributions. In contrast, larger datasets such as Diabetes, German, Sports, and Vehicle show relatively stable performance across a wider range of cluster numbers, suggesting that sufficient historical samples can mitigate the statistical uncertainty introduced by changes in context partitioning. In addition, multi-class and imbalanced datasets such as Accent and Zoo are also more sensitive to the number of clusters, possibly because minority-class samples are more easily diluted across different clusters, thereby affecting local reliability modeling. These observations suggest that the role of the cluster number is to balance contextual expressiveness and statistical stability: too few clusters may fail to capture reliability differences across sample regions, whereas too many clusters may amplify estimation fluctuations in small-sample or minority-class scenarios. A moderate clustering granularity is therefore generally more conducive to stable fusion performance.

\begin{figure*}[!t]
\centering
\includegraphics[max width=1.0\textwidth]{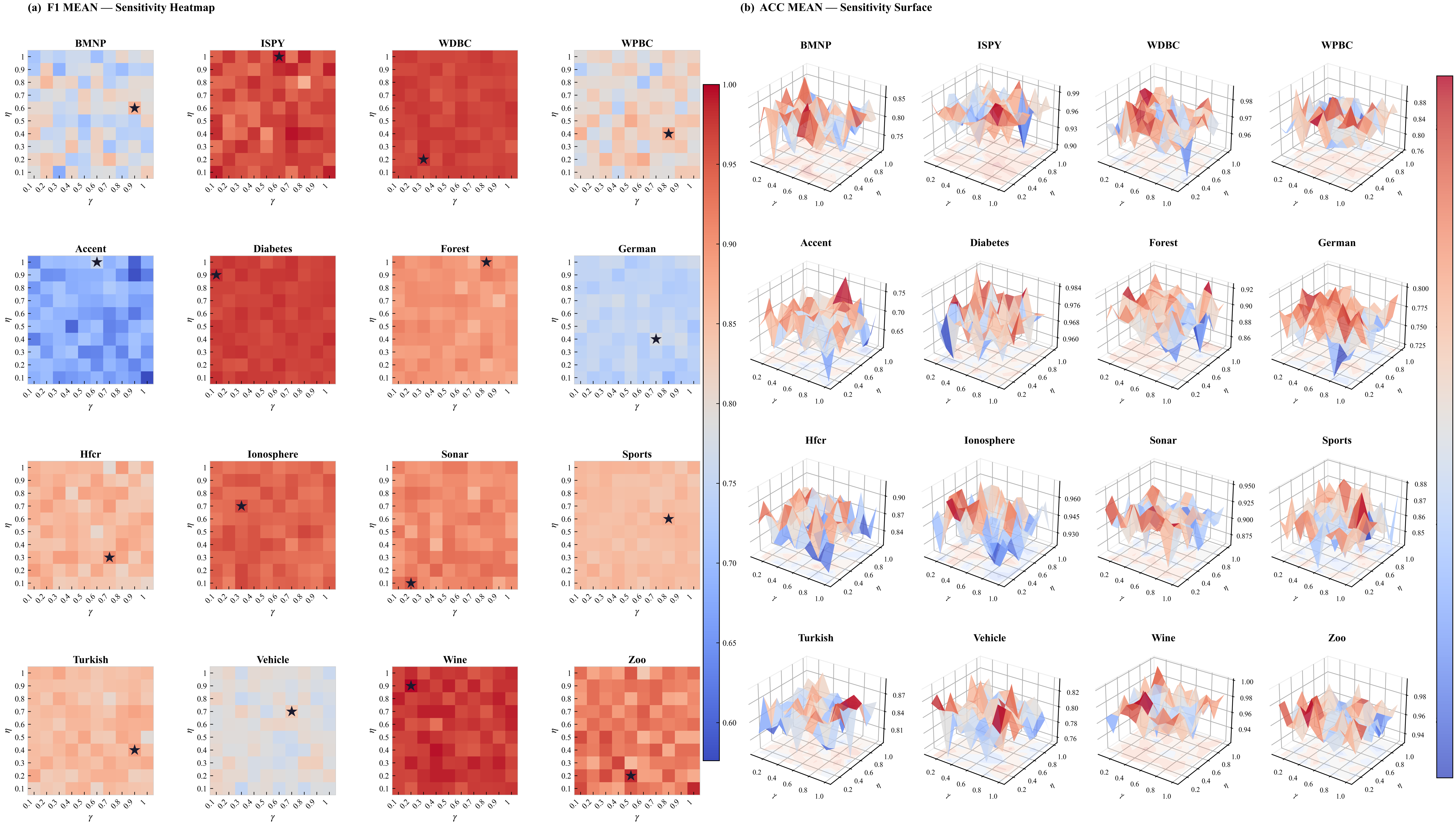}
\caption{Joint sensitivity analysis of hyperparameters \(\eta\) and \(\gamma\) across multiple datasets}
\label{fig:42_253_938_1291_738_0}
\end{figure*}

Figure~\ref{fig:42_253_938_1291_738_0} further reports the joint sensitivity results for \(\eta\) and \(\gamma\) . The F1 heatmaps on the left show that the optimal parameter combinations vary across datasets, indicating that the reward and penalty intensities should be adapted to the sample size, class distribution, and feature structure of each dataset. For datasets with relatively sufficient samples, such as WDBC, Diabetes, Sports, and Vehicle, high performance can be maintained over a relatively broad range of parameter combinations, forming continuous high-performance regions. This suggests that the historical experience weights in these scenarios are relatively tolerant to variations in \(\eta\) and \(\gamma\) . In contrast, small-sample datasets such as BMNP, Wine, WPBC, and Zoo exhibit more pronounced changes in the heatmaps, indicating that when historical samples are limited or class proportions are imbalanced, variations in the reward and penalty coefficients can more easily amplify local statistical errors in experience estimation and thus lead to performance fluctuations. For multi-class datasets such as Accent and Turkish, parameter sensitivity may also be jointly affected by inter-class separability and the reliability estimation of minority classes. The ACC surfaces on the right show a similar pattern: most datasets exhibit local fluctuations rather than a consistent increase along a single parameter direction, suggesting that excessively strong rewards or penalties may disrupt the balance of evidence weights. Overall, the effective regions of \(\eta\) and \(\gamma\) are usually not isolated points but are distributed across several neighboring parameter combinations, demonstrating a certain degree of parameter robustness in the proposed historical experience-driven mechanism. Meanwhile, the differences in optimal regions across datasets indicate that parameter selection should be moderately adjusted according to data scale, class imbalance, and feature complexity, so as to achieve a more appropriate balance between preserving historically reliable evidence and suppressing misleading evidence.

\subsubsection{Ablation study}

To examine the independent contribution of the key modules in the proposed framework, we conducted ablation experiments under the same experimental setting as that used in Section~\ref{sec:521}. The ablation variants include: replacing CCM with Dempster's conflict coefficient K; removing the clusters obtained by spectral clustering; separately removing the weights associated with rejoice and regret; completely removing historical experience weighting; retaining only the Dubois combination term or only the historical experience weighting term; and replacing the proposed decision rule with pignistic decision. All results are summarized in terms of the average F1 and AUC over the 16 datasets. The absolute and relative changes with respect to Full are also reported to characterize the influence of each component on classification performance and discriminative ability.

As shown in Table~\ref{tab:ablation_study_results_averaged_over_16_datasets}, the Full model achieves the best performance in both F1 and AUC. From the perspective of module contribution, historical experience weighting has the most pronounced impact. w/o History Weighting leads to a 3.21\% decrease in F1 and a 5.03\% decrease in AUC, producing the largest AUC loss among all variants. This suggests that historical feedback improves not only the final classification results but also the framework's overall ability to discriminate among evidence sources. In comparison, w/o Rejoice and w/o Regret result in F1 losses of 1.54\% and 1.98\%, respectively, indicating that both positive reinforcement and negative penalty contribute to reliability estimation. The larger effect of the regret term suggests that suppressing misleading evidence may be more critical in conflict-aware evidence fusion. w/o Cluster decreases F1 and AUC by 2.05\% and 1.57\%, respectively, further demonstrating the contextual dependency of evidence-source reliability. The degradation observed for w/ Dempster K indicates that the traditional conflict coefficient is insufficient to fully capture the internal uncertainty introduced by non-singleton focal elements, whereas the proposed CCM can more finely characterize both inter-evidence conflict and evidence non-specificity. Dubois Only suffers the largest F1 loss, suggesting that relying solely on conservative uncertainty preservation weakens the decisiveness of final classification. Although History Weighted Only can exploit historical reliability information, it still fails to achieve the stability of the full model without Dubois-type uncertainty preservation and conflict-adaptive regulation; consequently, its AUC degradation is larger than that of Dubois Only.

Overall, the ablation results support the core design logic of the proposed framework: CCM evaluates the conflict and uncertainty of the evidence set, historical experience weighting characterizes the long-term reliability of different evidence sources under similar contexts, and the hybrid fusion rule adaptively balances reliable consensus with uncertainty preservation. These components jointly improve the classification and discriminative performance of the model across heterogeneous datasets.

\begin{table*}[t]
\centering
\caption{Ablation study results averaged over 16 datasets}
\label{tab:ablation_study_results_averaged_over_16_datasets}
\adjustbox{max width=\textwidth}{
\begin{tabular}{ccccccc}
\toprule
Variant & F1 & \(\Delta \mathbf{F}\mathbf{1}\) (abs.) & \(\Delta \mathbf{F}\mathbf{1}\) (rel.) & AUC & $\Delta$AUC(abs.) & $\Delta$AUC(rel.)\\\cmidrule{1-7}
Full Model& \textbf{85.78±15.42} & - & - & \textbf{93.30±5.39} & - & - \\
\cmidrule{2-7}
w/ Pignistic&  84.62±15.63  &  -1.16  &  -1.35\%  &  92.91±5.83  &  -0.39  &  -0.42\% \\
w/ Dempster K&  84.31±16.03  &  -1.47  &  -1.71\%  &  92.14±5.75  &  -1.16  &  -1.24\% \\
\cmidrule{2-7}
w/o History Weighting&  83.03±15.56  &  -2.75  &  -3.21\%  &  88.61±5.87  &  -4.69  &  -5.03\% \\
w/o Cluster&  83.73±16.17  &  -2.05  &  -2.39\%  &  91.73±5.83  &  -1.57  &  -1.68\% \\
w/o Regret&  84.08±16.02  &  -1.70  &  -1.98\%  &  92.08±5.78  &  -1.22  &  -1.31\% \\
w/o Rejoice&  84.46±15.25  &  -1.32  &  -1.54\%  &  92.51±5.79  &  -0.79  &  -0.85\% \\
\cmidrule{2-7}
Dubois Only&  82.97±16.02  &  -2.81  &  -3.28\%  &  92.10±5.82  &  -1.20  &  -1.29\% \\
History Weighted Only&  83.69±16.20  &  -2.09  &  -2.44\%  &  91.36±5.72  &  -1.94  &  -2.08\% \\
\bottomrule
\end{tabular}
}
\end{table*}

\section{Conclusion}\label{sec:6_conclusion}

This paper has presented a unified evidence reasoning framework that combines CCM with historical-experience-driven weighting for robust multi-source decision making under uncertainty. The central premise of this work is that two persistent gaps in existing DST methods, namely the treatment of conflict and uncertainty as independent quantities and the neglect of long-term evidence source behavior, can be jointly addressed through complementary statistical and behavioral mechanisms. The CCM provides a unified scalar assessment of both inter-evidence inconsistency and intra-evidence non-specificity, grounded in a similarity measure whose mathematical properties are formally established. The historical experience weighting scheme exploits the observation that evidence sources operating repeatedly across heterogeneous decision contexts carry learnable reliability profiles, which RT makes it possible to quantify in terms of counterfactual regret and rejoice when fusion decisions deviate from ground truth. These two mechanisms are integrated through a hybrid combination rule that adaptively balances uncertainty preservation against weighted consensus, followed by a belief-interval decision strategy that produces deterministic classifications without discarding the epistemic uncertainty preserved by the fusion process.

The theoretical contributions of this work are threefold. First, the CCM offers a more faithful characterization of evidential stability than existing measures that examine conflict and uncertainty in isolation, and its five proven properties ensure consistent behavior across frame refinements and extreme cases. Second, the integration of SC with RT establishes a principled paradigm for translating historical decision feedback into context-dependent evidence weights, extending the scope of DST beyond instantaneous assessment toward adaptive reliability modeling. Third, the hybrid combination rule provides a conflict-aware mechanism for balancing the conservatism of fusion with the decisiveness of weighted consensus, parameterized by a single GCCD conflict indicator that requires no manual tuning of mixture coefficients.

On the practical side, experiments conducted across 16 heterogeneous datasets demonstrate that the proposed framework achieves competitive performance among both DST-based and ensemble learning methods, with the highest average F1 score (85.78) and mean AUC (93.30). The framework maintains its advantage under noisy training conditions, varying numbers of evidence sources, and different base evidence generators, indicating that the gains stem from the fusion architecture rather than from favorable data conditions. Ablation analysis confirms that historical experience weighting, CCM-based conflict assessment, and the hybrid combination rule each contribute meaningfully to performance, with the historical experience component exerting the strongest individual effect.

Several limitations suggest directions for further research. The current framework constructs BPAs through standard classifiers rather than dedicated evidence generation methods, and the quality of these assignments directly affects downstream fusion performance \citep{Zhou_2019}; developing domain-specific BPA generation strategies could improve both accuracy and computational efficiency. The computational cost of the CCM scales with the number of focal element pairs and may become prohibitive when the frame of discernment is large or the number of evidence sources is high \citep{Chen_2023}; approximate or distributed computation strategies would extend the framework's applicability to real-time decision settings. The sensitivity parameters governing the regret-rejoice mechanism, while demonstrating reasonable robustness across the datasets examined, may benefit from data-adaptive calibration rather than uniform specification. Future work should also explore the extension of the historical experience paradigm to heterogeneous multi-modal data environments, where evidence sources of fundamentally different types must be fused, and to nonstationary settings where the decision context itself evolves over time \citep{Strelet_2025,Zhang_2025}.

\appendix

\addcontentsline{toc}{section}{Appendix A}
\section*{\hypertarget{app:A}{Appendix A}}
\makeatletter
\@addtoreset{figure}{section}
\@addtoreset{table}{section}
\renewcommand{\thefigure}{\thesection.\arabic{figure}}
\renewcommand{\thetable}{\thesection.\arabic{table}}
\makeatother
\refstepcounter{section}
\onecolumn

\begin{table}[H]
\centering
\caption{Comparison of ACC on 16 datasets}
\small
\adjustbox{max width=\textwidth}{
\begin{tabular}{ccccccccccccc}
\toprule
Model & BHF & CatBoost & Dempster & Deng & Dubois & EC-FMDS & HEF & LightGBM & Murphy & Our & XGBoost & Yager\\\cmidrule{1-13}
Accent   &   69.01±3.95   &   60.79±4.36   &   67.47±4.33   &   69.00±2.00   &   65.96±1.64   &   65.95±2.31   &   62.00±3.52   & \textbf{72.94±3.47} &   66.57±6.11   &   69.05±12.42   &  71.41±5.50  &   60.82±6.98\\

BMNP    &    49.06±6.72    &    55.62±4.15    &    58.85±10.26    &    52.40±7.82    &    57.19±5.44    &    47.71±4.77    &    55.73±10.67    &    66.67±5.89    &    54.27±17.33    & \textbf{71.35±34.12} &    61.77±8.15    &    52.40±7.82\\

Diabetes    &    95.77±0.77    &    93.08±2.26    &    93.85±1.88    &    95.58±3.40    &    96.73±1.46    &    97.31±0.77    &    96.73±1.71    &    91.15±1.47    &    95.00±0.99    & \textbf{97.69±1.26} &    94.23±2.55    &    95.96±0.74\\

Forest    &    84.89±2.93    &    86.22±3.31    &    86.23±4.71    &    86.03±2.94    &    87.56±2.50    &    85.46±3.19    &    82.98±4.40    &    87.75±4.89    &    85.46±4.34    & \textbf{88.72±5.11} &    88.14±3.59    &    84.71±1.36\\

German    &    66.60±1.80    &    73.20±2.38    &    69.50±4.12    &    68.70±1.74    &    70.10±2.76    &    69.20±4.88    &    68.60±2.58    &    74.40±2.19    &    70.00±2.79    & \textbf{74.80±3.68} &    74.00±4.36    &    68.20±2.39\\

Hfcr    &    79.95±3.85    &    79.59±3.89    &    82.96±7.38    &    81.29±4.80    &    80.94±7.79    &    83.63±5.85    &    79.60±7.25    & \textbf{84.96±7.54} &    81.95±6.16    &    84.29±4.22    &    80.95±8.50    &    82.62±4.70\\

Ionosphere    &    89.18±4.66    &    89.46±3.28    &    90.33±5.03    &    91.75±2.97    &    88.90±3.73    &    91.75±4.84    &    90.03±3.76    &    91.46±3.51    &    93.17±3.33    & \textbf{94.03±5.44} &    89.76±4.62    &    89.47±6.02\\

ISPY    &    96.73±1.82    & \textbf{99.07±1.08} &    95.80±2.83    &    95.35±1.84    &    94.40±3.45    &    92.49±10.24    &    92.57±5.44    & \textbf{99.07±1.08} &    97.67±1.79    &    94.38±5.14    & \textbf{99.07±1.08} &    90.23±3.81\\

Sonar    &    71.63±1.84    &    73.08±4.15    &    68.75±5.06    &    77.40±4.54    &    74.04±4.00    &    75.96±7.11    &    70.67±5.95    &    77.88±5.98    &    76.92±6.84    & \textbf{90.87±12.10} &    75.48±3.28    &    75.48±6.35\\

Sports    &    77.70±3.24    &    80.30±1.71    &    78.30±2.68    &    77.20±1.18    &    77.20±3.12    &    79.30±2.78    &    77.10±2.43    &    80.50±3.17    &    80.30±3.88    & \textbf{87.70±7.14} &    79.20±1.57    &    74.80±3.41\\

Turkish    &    67.25±2.22    &    65.50±4.04    &    67.75±5.85    &    67.25±3.59    &    66.25±4.11    &    67.75±3.40    &    63.50±5.80    &    78.25±2.50    &    69.25±0.96    & \textbf{86.00±14.21} &    74.50±4.80    &    66.50±7.59\\

Vehicle    &    73.41±2.80    &    68.56±1.17    &    69.74±1.52    &    72.82±1.35    &    70.93±4.20    &    74.00±1.32    &    68.20±2.39    &    72.93±2.02    &    73.88±2.64    & \textbf{77.91±7.89} &    73.41±1.88    &    69.38±2.51\\

WDBC    &    95.25±1.56    &    94.55±1.84    &    93.67±1.73    &    95.43±1.35    &    93.15±1.04    &    94.73±1.67    &    93.67±2.64    &    95.08±2.99    &    94.90±2.33    & \textbf{98.42±1.05} &    94.03±3.06    &    95.08±0.99\\

Wine    &    93.23±4.93    &    94.95±3.86    &    93.80±5.00    &    95.52±3.18    &    93.24±3.19    &    94.37±6.01    &    89.87±4.37    &    92.68±3.88    &    96.06±3.88    & \textbf{98.89±2.22} &    93.27±3.17    &    88.79±4.45\\

WPBC    &    68.71±6.12    &    79.29±4.55    &    73.74±4.33    &    68.17±7.87    &    73.23±4.14    &    73.24±4.36    &    73.73±3.72    &    75.78±5.10    &    66.70±6.45    & \textbf{80.89±14.21} &    74.78±3.96    &    73.23±1.88\\

Zoo    &    91.12±6.77    &    90.04±9.55    &    94.04±6.94    &    93.15±6.56    &    93.08±1.95    &    92.12±7.22    &    91.12±3.71    &    88.15±4.45    &    92.12±5.55    & \textbf{97.12±5.77} &    92.12±4.49    &    93.15±8.03\\

\bottomrule
\end{tabular}
}
\end{table}

\begin{table}[H]
\centering
\caption{Comparison of PRE on 16 datasets}
\small
\adjustbox{max width=\textwidth}{
\begin{tabular}{ccccccccccccc}
\toprule
Model & BHF & CatBoost & Dempster & Deng & Dubois & EC-FMDS & HEF & LightGBM & Murphy & Our & XGBoost & Yager\\\cmidrule{1-13}
Accent   &   62.90±3.39   &   60.51±13.33   &   68.26±4.36   &   63.51±3.49   &   60.13±2.21   &   61.48±3.65   &   63.55±4.92   & \textbf{68.89±8.55} &   60.92±7.25   &   66.04±11.57   &  68.75±4.72  &   54.61±8.64\\

BMNP    &    34.28±7.75    &    34.31±4.61    &    36.71±23.19    &    45.16±10.35    &    43.11±12.96    &    42.34±11.65    &    37.67±15.76    &    39.06±6.48    &    51.11±24.70    & \textbf{68.65±36.95} &    45.33±12.11    &    41.44±10.14\\

Diabetes    &    95.30±0.70    &    92.53±2.56    &    93.21±2.03    &    95.31±3.84    &    96.31±1.72    &    97.04±0.91    &    96.32±1.97    &    91.00±1.90    &    94.50±1.19    & \textbf{97.38±1.56} &    93.80±2.81    &    95.48±0.94\\

Forest    &    84.95±1.90    &    86.59±2.94    &    86.71±4.58    &    86.30±1.56    &    87.63±0.91    &    85.38±2.55    &    85.36±4.41    &    88.39±3.66    &    84.88±3.64    & \textbf{89.07±5.92} &    88.67±2.77    &    85.00±2.77\\

German    &    62.54±3.11    &    67.56±3.83    &    64.73±4.11    &    63.09±2.58    &    65.24±3.25    &    64.26±4.79    &    63.51±3.25    &    69.15±3.25    &    65.38±2.90    & \textbf{71.69±3.81} &    68.74±5.56    &    63.63±2.55\\

Hfcr    &    77.46±4.67    &    77.76±4.20    &    80.63±8.30    &    78.99±5.00    &    78.04±8.99    &    81.13±6.62    &    76.92±8.25    & \textbf{82.96±8.59} &    79.12±6.88    &    82.32±4.50    &    78.34±9.80    &    80.61±4.83\\

Ionosphere    &    89.07±4.99    &    89.76±3.52    &    89.52±5.46    &    90.85±2.64    &    88.11±4.05    &    91.21±5.14    &    88.77±3.78    &    91.88±4.10    &    93.05±3.37    & \textbf{93.52±6.04} &    90.16±5.40    &    89.07±6.16\\

ISPY    &    95.35±3.03    & \textbf{99.38±0.72} &    96.01±2.76    &    94.55±1.37    &    93.56±4.54    &    92.03±12.77    &    93.43±6.98    & \textbf{99.38±0.72} &    97.00±2.82    &    92.06±5.94    & \textbf{99.38±0.72} &    88.67±3.58\\

Sonar    &    71.58±1.91    &    74.12±2.66    &    70.23±5.61    &    77.40±4.55    &    74.58±3.34    &    76.37±7.39    &    71.77±7.24    &    78.24±6.35    &    77.05±6.96    & \textbf{91.00±12.12} &    75.46±3.31    &    75.47±6.28\\

Sports    &    75.98±3.43    &    79.38±2.17    &    77.42±2.88    &    75.58±1.26    &    75.80±3.66    &    77.77±2.98    &    75.49±2.79    &    79.27±3.30    &    78.76±4.19    & \textbf{86.68±7.64} &    77.81±1.93    &    72.90±3.79\\

Turkish    &    68.04±3.38    &    66.44±4.37    &    68.36±6.03    &    67.25±3.99    &    67.33±3.03    &    69.78±2.62    &    66.48±4.82    &    79.41±2.24    &    69.63±1.00    & \textbf{86.04±14.46} &    74.80±4.44    &    69.87±7.70\\

Vehicle    &    73.11±3.37    &    66.73±0.72    &    68.92±2.02    &    72.32±1.46    &    70.50±4.96    &    73.56±1.70    &    67.43±2.83    &    71.88±2.65    &    73.23±2.77    & \textbf{77.42±8.69} &    73.03±1.88    &    68.61±2.48\\

WDBC    &    94.92±1.80    &    94.72±1.95    &    93.00±1.99    &    95.08±1.59    &    92.46±1.22    &    94.25±1.90    &    93.04±3.02    &    94.92±3.66    &    94.83±2.78    & \textbf{98.12±1.11} &    93.45±3.35    &    94.48±1.02\\

Wine    &    93.35±4.99    &    95.32±3.82    &    94.71±3.90    &    95.54±3.16    &    93.82±2.78    &    94.69±5.67    &    91.23±3.67    &    93.30±4.01    &    96.18±3.74    & \textbf{98.90±2.21} &    93.72±3.34    &    89.59±4.42\\

WPBC    &    55.97±8.11    &    70.35±10.70    &    60.78±7.40    &    55.68±9.47    &    56.80±10.59    &    60.98±8.46    &    58.91±8.63    &    66.32±8.17    &    56.10±6.07    & \textbf{75.24±16.54} &    64.07±5.59    &    57.90±7.36\\

Zoo    &    79.46±14.33    &    82.29±12.16    &    89.88±14.07    &    83.33±16.64    &    82.14±9.86    &    78.51±16.32    &    75.95±3.05    &    71.21±10.93    &    81.85±12.35    & \textbf{92.86±14.29} &    80.89±9.89    &    83.04±22.09\\

\bottomrule
\end{tabular}
}
\end{table}

\begin{table}[H]
\centering
\caption{Comparison of REC on 16 datasets}
\small
\adjustbox{max width=\textwidth}{
\begin{tabular}{ccccccccccccc}
\toprule
Model & BHF & CatBoost & Dempster & Deng & Dubois & EC-FMDS & HEF & LightGBM & Murphy & Our & XGBoost & Yager\\\cmidrule{1-13}
Accent   &  65.16±5.77  &   41.79±7.27   &   61.05±8.66   &   63.24±6.96   &   62.17±5.32   &   62.86±4.20   &   52.56±7.72   &   62.17±4.57   &   59.75±6.52   & \textbf{76.71±16.43} &   63.38±7.60   &   56.99±7.94\\

BMNP    &    37.96±7.22    &    38.15±3.79    &    39.49±13.90    &    44.31±2.37    &    44.72±11.59    &    40.56±6.48    &    40.32±11.50    &    43.24±5.86    &    44.49±19.53    & \textbf{72.87±40.50} &    45.88±9.22    &    42.22±9.65\\

Diabetes    &    96.00±1.13    &    93.16±2.21    &    94.44±1.47    &    95.66±3.00    &    97.06±1.05    &    97.34±0.77    &    96.97±1.47    &    90.47±0.99    &    95.28±0.57    & \textbf{97.84±1.04} &    94.38±2.07    &    96.25±0.62\\

Forest    &    83.73±5.00    &    85.46±5.14    &    84.92±4.85    &    84.91±4.69    &    86.66±4.24    &    85.28±5.26    &    81.59±5.29    &    86.66±6.19    &    85.48±5.93    & \textbf{89.48±4.37} &    86.44±4.89    &    84.98±0.91\\

German    &    64.05±4.07    &    62.86±3.00    &    65.64±3.91    &    63.55±3.26    &    66.26±3.61    &    64.76±5.10    &    64.43±3.80    &    65.05±3.93    &    66.57±3.08    & \textbf{74.67±4.54} &    66.57±5.47    &    64.81±3.20\\

Hfcr    &    80.01±6.28    &    73.17±5.71    &    79.20±10.48    &    81.01±4.07    &    77.45±9.89    &    82.46±7.00    &    75.38±9.34    &    81.79±9.62    &    80.12±7.78    & \textbf{83.77±2.84} &    76.91±10.66    &    78.69±7.58\\

Ionosphere    &    87.24±5.37    &    87.27±3.90    &    90.60±5.10    &    91.50±4.13    &    88.60±3.98    &    90.99±5.66    &    90.33±4.66    &    89.71±3.87    &    92.09±3.90    & \textbf{93.60±5.68} &    87.67±5.00    &    88.84±5.73\\

ISPY    &    96.69±2.06    & \textbf{98.27±2.00} &    93.36±5.16    &    93.61±3.26    &    92.97±5.18    &    90.98±9.41    &    87.17±8.53    & \textbf{98.27±2.00} &    97.31±2.03    &    96.16±3.48    & \textbf{98.27±2.00} &    86.32±7.08\\

Sonar    &    71.52±1.85    &    72.35±5.00    &    69.45±5.06    &    77.14±4.76    &    74.27±3.81    &    75.86±7.06    &    71.13±6.35    &    77.59±5.89    &    77.10±6.96    & \textbf{90.84±12.01} &    75.34±3.31    &    75.42±6.28\\

Sports    &    76.57±4.02    &    77.33±1.63    &    74.59±3.48    &    75.30±2.49    &    74.37±3.53    &    77.54±3.28    &    74.52±2.59    &    78.14±4.04    &    78.72±4.32    & \textbf{87.05±7.43} &    76.87±1.44    &    71.72±4.03\\

Turkish    &    67.25±2.22    &    65.50±4.04    &    67.75±5.85    &    67.25±3.59    &    66.25±4.11    &    67.75±3.40    &    63.50±5.80    &    78.25±2.50    &    69.25±0.96    & \textbf{86.00±14.21} &    74.50±4.80    &    66.50±7.59\\

Vehicle    &    73.70±2.82    &    68.81±1.17    &    70.03±1.66    &    73.14±1.48    &    71.21±4.16    &    74.31±1.41    &    68.63±2.37    &    73.26±2.05    &    74.13±2.70    & \textbf{78.06±8.00} &    73.68±1.94    &    69.77±2.50\\

WDBC    &    94.97±1.52    &    93.65±2.13    &    93.71±1.57    &    95.21±1.25    &    93.10±0.96    &    94.55±1.64    &    93.71±2.28    &    94.74±2.48    &    94.31±2.13    & \textbf{98.55±1.16} &    93.99±2.92    &    95.22±1.17\\

Wine    &    93.73±4.53    &    95.24±3.45    &    93.49±5.80    &    96.11±2.71    &    93.73±3.00    &    94.78±5.31    &    90.83±3.37    &    93.06±3.50    &    96.20±3.32    & \textbf{98.98±2.04} &    93.53±2.47    &    88.87±4.46\\

WPBC    &    55.36±6.70    &    63.12±10.62    &    57.95±5.95    &    55.00±8.74    &    55.50±7.72    &    60.76±9.73    &    57.24±7.37    &    60.09±5.80    &    56.26±5.99    & \textbf{80.93±20.04} &    62.48±7.71    &    55.23±5.76\\

Zoo    &    84.88±10.30    &    82.74±9.76    &    90.12±10.78    &    85.60±14.13    &    85.36±9.43    &    83.93±12.20    &    81.43±3.50    &    75.60±4.91    &    82.74±12.20    & \textbf{94.64±10.71} &    85.71±5.83    &    85.36±17.70\\

\bottomrule
\end{tabular}
}
\end{table}

\begin{table}[H]
\centering
\caption{Comparison of F1 on 16 datasets}
\small
\adjustbox{max width=\textwidth}{
\begin{tabular}{ccccccccccccc}
\toprule
Model & BHF & CatBoost & Dempster & Deng & Dubois & EC-FMDS & HEF & LightGBM & Murphy & Our & XGBoost & Yager\\\cmidrule{1-13}
Accent   &   62.90±4.48   &   44.45±7.66   &   59.40±7.33   &   61.94±4.29   &   59.94±3.02   &   60.81±1.79   &   49.54±7.05   &   63.15±6.40   &   59.06±6.93   & \textbf{68.33±14.48} &  64.44±6.79  &   54.58±8.38\\

BMNP    &    35.48±7.32    &    36.06±4.25    &    35.30±14.95    &    43.54±4.79    &    42.77±11.46    &    40.05±7.24    &    37.91±13.18    &    40.53±6.06    &    44.73±19.78    & \textbf{68.34±38.34} &    44.39±10.32    &    39.91±8.39\\

Diabetes    &    95.57±0.82    &    92.76±2.36    &    93.62±1.90    &    95.38±3.52    &    96.59±1.48    &    97.17±0.81    &    96.58±1.76    &    90.62±1.42    &    94.78±0.98    & \textbf{97.58±1.31} &    93.98±2.59    &    95.78±0.74\\

Forest    &    83.90±3.72    &    85.80±4.27    &    85.59±4.75    &    85.22±3.45    &    86.66±3.17    &    85.00±4.00    &    82.50±4.55    &    87.20±5.32    &    84.93±4.93    & \textbf{88.77±5.44} &    87.25±4.04    &    84.57±1.26\\

German    &    62.62±3.01    &    63.75±3.39    &    65.02±4.14    &    63.25±2.84    &    65.59±3.36    &    64.25±4.90    &    63.82±3.41    &    65.99±4.22    &    65.75±3.00    & \textbf{72.23±4.09} &    67.24±5.58    &    63.85±2.69\\

Hfcr    &    78.03±4.83    &    74.49±5.52    &    79.55±9.67    &    79.52±4.83    &    77.68±9.49    &    81.62±6.68    &    75.79±8.70    &    82.25±9.16    &    79.55±7.25    & \textbf{82.63±4.09} &    77.38±10.31    &    79.13±6.82\\

Ionosphere    &    87.97±5.20    &    88.19±3.77    &    89.73±5.32    &    91.06±3.39    &    88.08±3.87    &    91.01±5.27    &    89.37±4.07    &    90.52±3.85    &    92.50±3.66    & \textbf{93.53±5.87} &    88.58±5.10    &    88.67±6.19\\

ISPY    &    95.91±2.30    & \textbf{98.79±1.40} &    94.42±3.89    &    94.02±2.44    &    92.85±4.33    &    91.03±11.58    &    89.56±8.02    & \textbf{98.79±1.40} &    97.09±2.23    &    93.46±5.68    & \textbf{98.79±1.40} &    87.07±5.50\\

Sonar    &    71.49±1.82    &    72.05±5.51    &    68.61±4.97    &    77.18±4.74    &    73.90±4.10    &    75.77±7.04    &    70.55±5.92    &    77.63±6.03    &    76.89±6.89    & \textbf{90.85±12.09} &    75.33±3.32    &    75.37±6.34\\

Sports    &    76.19±3.65    &    78.07±1.77    &    75.42±3.39    &    75.27±1.80    &    74.79±3.50    &    77.58±3.12    &    74.87±2.61    &    78.53±3.78    &    78.73±4.23    & \textbf{86.84±7.56} &    77.23±1.56    &    72.11±3.99\\

Turkish    &    66.65±1.90    &    65.42±4.12    &    66.94±6.13    &    66.75±3.81    &    66.14±4.06    &    67.20±3.34    &    63.61±5.73    &    78.19±2.30    &    69.10±1.14    & \textbf{85.92±14.38} &    74.28±4.40    &    66.90±7.31\\

Vehicle    &    73.23±3.20    &    67.23±0.86    &    68.99±1.77    &    72.51±1.58    &    70.68±4.62    &    73.77±1.53    &    66.68±2.45    &    72.34±2.42    &    73.54±2.80    & \textbf{77.22±8.35} &    73.18±1.90    &    68.57±2.84\\

WDBC    &    94.93±1.64    &    94.10±2.01    &    93.30±1.79    &    95.13±1.42    &    92.73±1.08    &    94.39±1.76    &    93.31±2.72    &    94.77±3.10    &    94.53±2.44    & \textbf{98.32±1.12} &    93.67±3.19    &    94.78±1.05\\

Wine    &    93.28±4.98    &    95.08±3.76    &    93.62±5.32    &    95.63±3.10    &    93.48±3.00    &    94.38±5.90    &    90.46±3.99    &    92.71±3.83    &    95.87±3.85    & \textbf{98.92±2.17} &    93.35±2.97    &    88.67±4.58\\

WPBC    &    55.41±7.19    &    63.27±12.49    &    58.14±6.87    &    55.27±9.03    &    55.01±9.58    &    60.38±9.13    &    57.02±8.81    &    60.67±6.64    &    56.06±6.16    & \textbf{76.62±17.67} &    62.51±6.72    &    54.69±6.52\\

Zoo    &    80.38±13.36    &    81.28±11.22    &    88.86±12.64    &    83.82±15.59    &    82.28±10.43    &    79.78±15.17    &    76.57±3.18    &    71.99±8.23    &    80.71±13.46    & \textbf{92.86±14.29} &    82.01±7.35    &    83.23±20.70\\

\bottomrule
\end{tabular}
}
\end{table}

\addcontentsline{toc}{section}{Appendix B}
\section*{\hypertarget{app:B}{Appendix B}}
\refstepcounter{section}

\begin{figure}[H]
\centering
\includegraphics[width=0.7\linewidth]{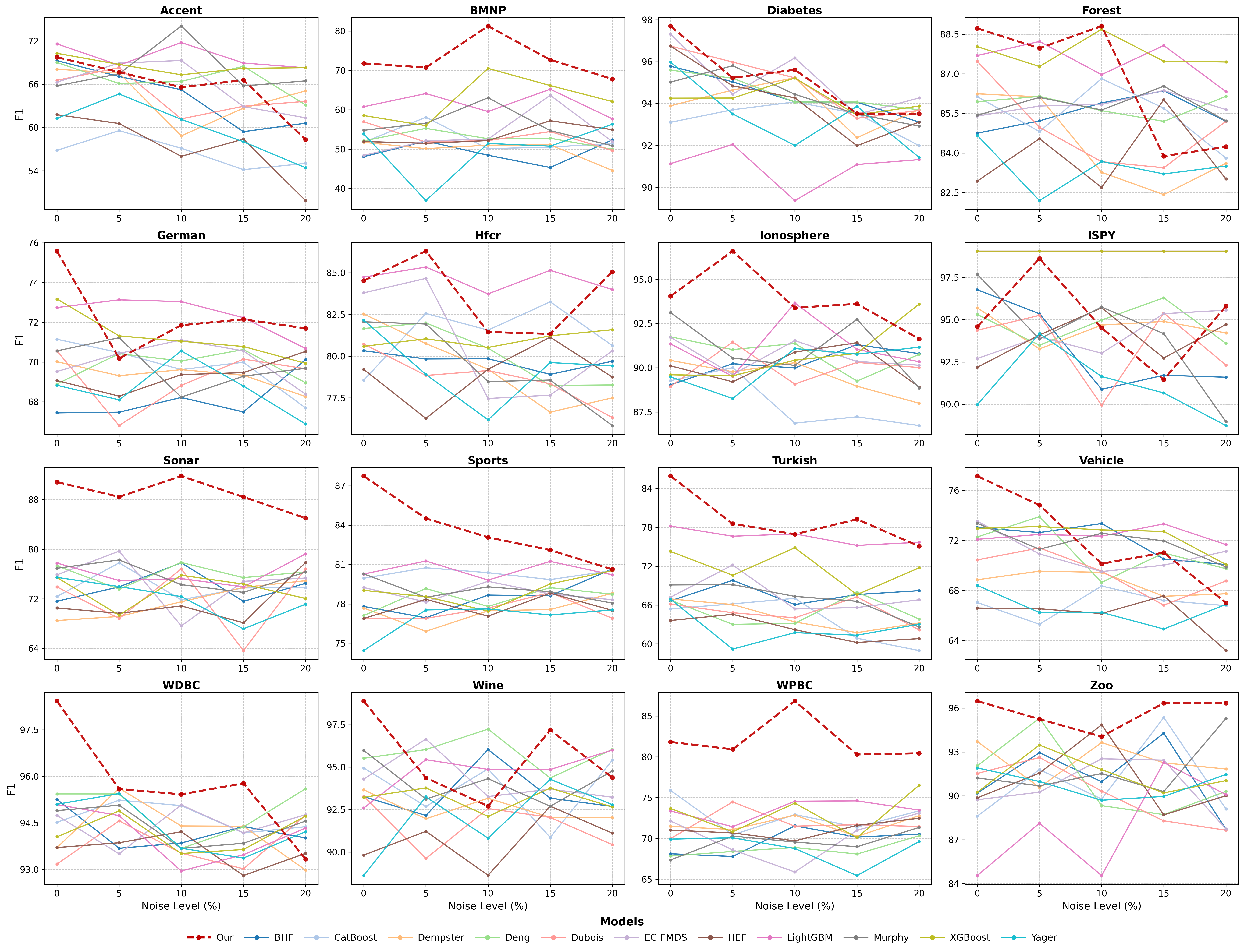}
\caption{F1-score under Gaussian feature noise}
\label{fig:47_320_739_1164_888_0}
\end{figure}

\begin{figure}[H]
\centering
\includegraphics[width=0.7\linewidth]{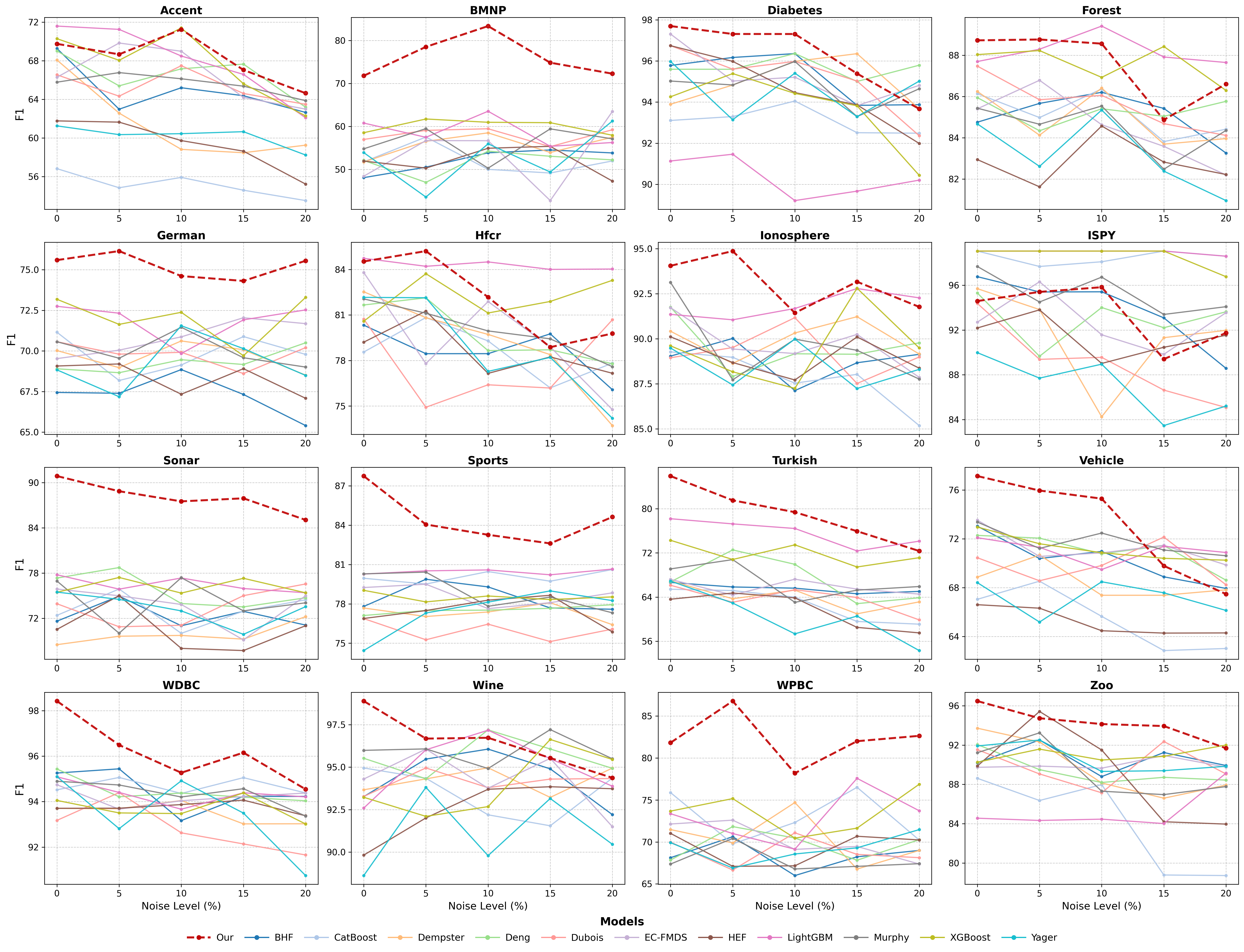}
\caption{F1-score under mean feature noise}
\label{fig:48_319_196_1163_895_0}
\end{figure}

\begin{figure}[H]
\centering
\includegraphics[width=0.7\linewidth]{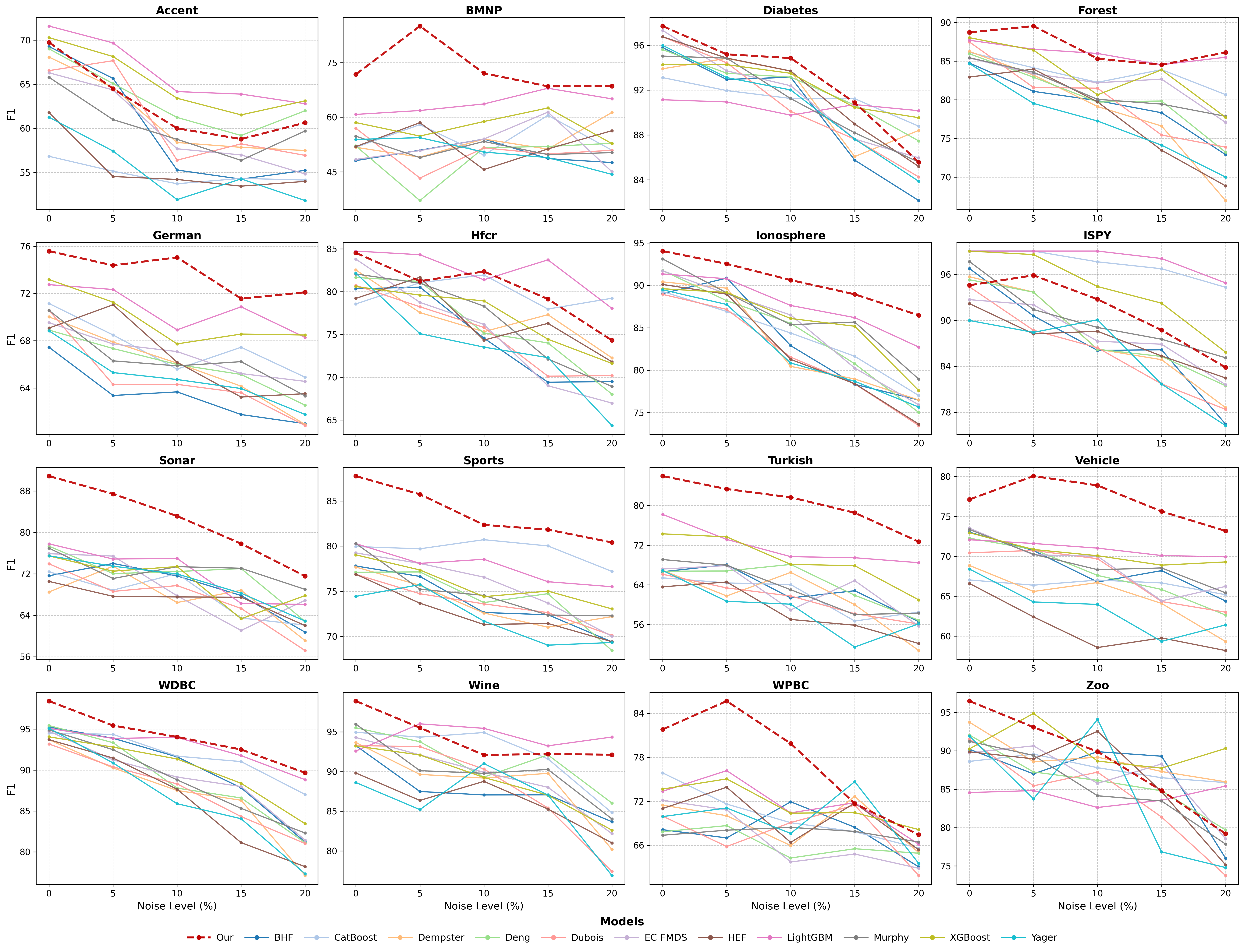}
\caption{F1-score under random label noise}
\label{fig:48_320_1154_1161_880_0}
\end{figure}

\addcontentsline{toc}{section}{References}
\printbibliography

\end{document}